\documentclass[10pt,twocolumn,letterpaper]{article}
\usepackage{graphicx}
\usepackage{amsmath}
\usepackage{xspace}
\usepackage{tabularx}
\usepackage{booktabs}
\usepackage{multicol}
\usepackage{multirow}
\usepackage{color}
\usepackage{adjustbox}
\usepackage{graphicx}
\usepackage{subcaption}
\usepackage{pgf-pie}
\usepackage[table]{xcolor}
\usepackage{wrapfig}
\usepackage{tikz}
\usepackage[most]{tcolorbox}
\usepackage{alltt}

\tcbset{
  aibox/.style={
    width=\linewidth,
    top=0pt,
    colback=white,
    colframe=black,
    colbacktitle=black,
    enhanced,
    center,
    attach boxed title to top left={yshift=-0.1in,xshift=0.1in},
    boxed title style={boxrule=0pt,colframe=white,},
  }
}

\newtcolorbox{AIbox}[2][]{aibox,title=#2,#1}

\newcommand{\takeawaybox}[1]{%
  \begin{tcolorbox}[
    colback=gray!20!white,
    boxrule=0pt,
    arc=0mm,
    left=5pt,
    right=5pt,
    top=2pt,
    bottom=2pt,
    fontupper={\fontsize{10pt}{11.75pt}\selectfont}
  ]
    \textbf{Takeaway:} #1
  \end{tcolorbox}
  \vspace*{-0.1cm}
}

\usepackage{xspace}
\newcommand{\ours}{OctoMed\xspace}

\usepackage[pagenumbers]{cvpr}

\definecolor{cvprblue}{rgb}{0.21,0.49,0.74}
\usepackage[pagebackref,breaklinks,colorlinks,allcolors=cvprblue]{hyperref}

\usepackage{FiraSans}
\newcommand{\octomedtitle}{
  \includegraphics[width=20pt]{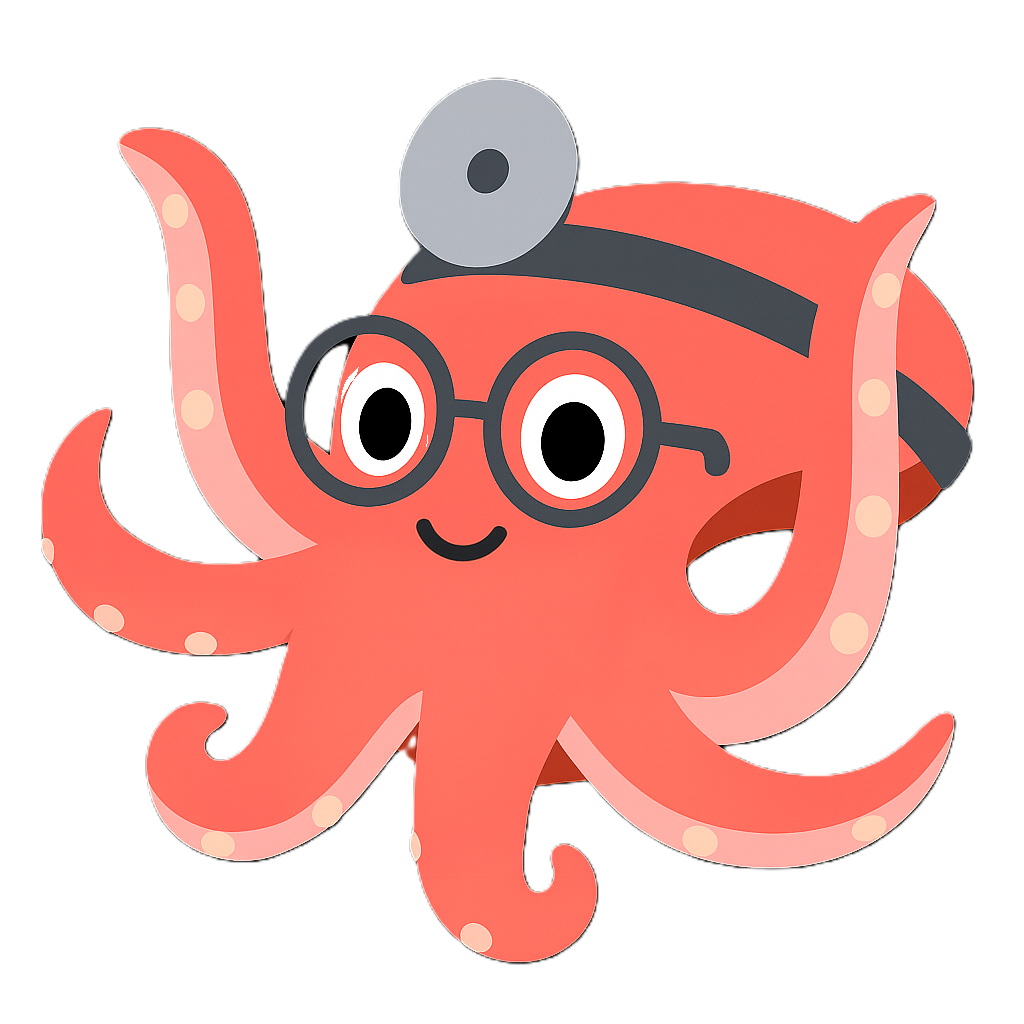}{\fontfamily{FiraSans-TLF}\selectfont OctoMed}\xspace%
}

\title{\octomedtitle: Data Recipes for State-of-the-Art Multimodal Medical Reasoning}

\author{
  Timothy Ossowski$^{1,}$\thanks{\scriptsize Equal contribution} \quad
  Sheng Zhang$^{2,}$\footnotemark[1] \quad \\
  Qianchu Liu$^{2}$\quad
  Guanghui Qin$^{2}$\quad
  Reuben Tan$^{2}$\quad 
  Tristan Naumann$^{2}$\quad
  Junjie Hu$^{1}$\quad
  Hoifung Poon$^{2}$\\[6pt]
  {\small $^{1}$University of Wisconsin–Madison, WI, USA} \quad 
  {\small $^{2}$Microsoft Research, Redmond, WA, USA}
}

\begin{document}
\maketitle
\begin{abstract}
High-quality and carefully curated data is a cornerstone of training medical large language models, as it directly impacts both generalization and robustness to unseen clinical tasks. We investigate strategies for training and data curation to develop a robust multimodal reasoning model in the medical domain. Our work focuses on supervised fine-tuning (SFT) and explores data recipes that leverage structured reasoning traces. Using our proposed data recipe, we scale experiments to a dataset of over 8 million examples and 6.8 billion response tokens, achieving state-of-the-art performance among open-source models across diverse out-of-distribution medical benchmark tasks. Our results further indicate that curating a high-quality, diverse training dataset with varying structured reasoning trace lengths enables the fine-tuned model to self-calibrate its reasoning trajectory lengths based on the downstream task, without explicit supervision. We present key insights, describe the data curation strategy, and outline next steps toward developing robust medical vision-language reasoning system\footnote{\scriptsize Model URL: \url{https://huggingface.co/OctoMed/OctoMed-7B}}.
\end{abstract}    
\begin{figure*}[!t]
    \centering
    \includegraphics[width=\linewidth]{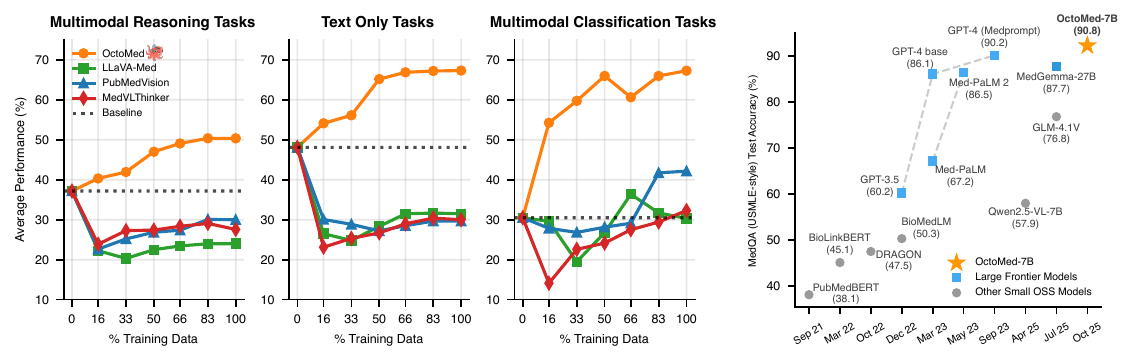}
    \caption{\textbf{Left:} Average performance on 3 task types when finetuning a student model with various SFT datasets. All student models were initialized with the same \texttt{Qwen2.5-VL-7B-Instruct} checkpoint and compared to the student's  performance before finetuning (dotted line). \textbf{Right: } Progress on MedQA performance over time. Despite its modest 7B parameter size, \ours outperforms strong open small-scale and large proprietary systems.}
    \label{fig:performance_over_time}
\end{figure*}

\section{Introduction}

Recent advances in large language models (LLMs) and multimodal reasoning systems have produced striking results across open-domain tasks such as general question answering, planning, and multi-step reasoning. These models often start from strong general-purpose backbones and gain reasoning capabilities through carefully designed post-training pipelines, such as supervised fine-tuning (SFT) or reinforcement learning (RL). However, applying such pipelines to medical reasoning presents a unique challenge that differs substantially from open-domain settings. Medical reasoning must integrate heterogeneous signals across diverse, high-stakes modalities (e.g., radiology images \cite{zambranochaves2024llavarad, Bannur2024MAIRA2GR}, granular pathology slides \cite{zhang2025patho, xu2024whole, lu2023foundational}, structured lab values, and complex clinical notes \cite{rasmy2020medbert, wang2023hierarchical, huang2024heart, adibvafa2024ehrmamba}), interpret noisy or unseen observations, and support long-horizon, safety-critical decisions. Consequently, multimodal medical reasoning demands models that can operate under substantial data distribution shifts, integrate information across a wide range of modalities, and maintain levels of fidelity and robustness far beyond those required in typical open-domain deployment.

A central, yet often underexplored, design factor in achieving medical reasoning robustness lies not in architectural novelty or larger model backbones, but in the \textit{effective curation of the data used to teach models how to reason}. The mixture of questions, modalities, and reasoning traces a model sees during SFT strongly shapes its generalization ability and calibration. 
A narrow or imbalanced mixture can easily yield overfitted models that fail on out-of-distribution cases, a finding explored both theoretically \cite{vardhan2025learning} and empirically \cite{liu2024regmix} in predictive modeling. Furthermore, recent findings~\cite{yue2025does} suggest that current RL paradigms largely refine behaviors already acquired in SFT rather than introduce fundamentally new reasoning patterns. This motivates a thorough, data-centric examination of how multimodal medical reasoning data should be structured, scaled, and balanced to most effectively enhance reasoning during SFT.

\begin{figure*}[t]
    \centering
    \includegraphics[width=1.0\linewidth]{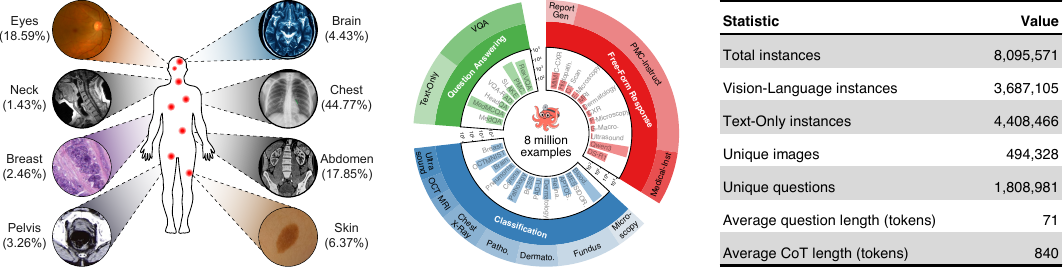}
    \caption{Overview of the SFT dataset. \textbf{Left:} Distribution of imaging modalities and anatomical regions represented in the SFT mixture. For large datasets in our mixture lacking modality and region annotations (e.g., PMC-VQA), we obtained this metadata by prompting \texttt{GPT-4.1-mini}. The percentages do not total 100\% due to a minor fraction of samples from other less common modalities. \textbf{Middle:} Breakdown of task types and source datasets used for distillation. \textbf{Right:} Summary of key dataset statistics.}
    \label{fig:data_dist}
\end{figure*}

This work places data curation at the center of developing medical reasoning models, and asks:
\textit{What recipes for structuring, balancing, and scaling multimodal reasoning data most effectively improve medical reasoning?} We propose a structured data recipe for multimodal medical reasoning that leverages SFT while emphasizing diversity in reasoning trace lengths and coverage across modalities and task types. Using this recipe, we construct the largest medical multimodal reasoning dataset, containing over 8 million examples (Figure \ref{fig:data_dist}), spanning both text and image data. Each reasoning trace is carefully curated for quality via rejection sampling, ensuring high-fidelity reasoning paths. We further provide key insights into strategies for developing robust multimodal medical reasoning models. We summarize our key findings and contributions below.
\begin{itemize}
    \item \textbf{State-of-the-Art Performance:} Using our curated data recipe, we develop \ours, achieving state-of-the-art results on diverse medical reasoning benchmarks (Figure \ref{fig:performance_over_time}; Table \ref{tab:performances}). Notably, our model remains competitive on reasoning-intensive text-only tasks, where many existing multimodal models underperform.
    \item \textbf{Scaling and Diversifying Reasoning Traces:} We show that incorporating multiple valid reasoning traces per example and expanding modality coverage proves more effective than simply increasing training epochs. We identify key data curation strategies and scale our recipe to obtain the new largest medical reasoning dataset.
    \item \textbf{Emergent Task-Aware Reasoning:} Training on a mixture of varying reasoning trace lengths leads to dynamic adaptation of reasoning depth to task complexity. Without explicit supervision, \ours produces longer and more detailed reasoning traces on challenging or out-of-distribution benchmarks, revealing an interpretable signal of task difficulty that could guide future post-training or data-filtering pipelines.
\end{itemize}

\section{Methodology}

We adopt a similar approach to existing SFT post-trained reasoning models, leveraging knowledge distillation from a strong teacher model $\mathcal{T}$, such as \texttt{DeepSeek-R1}. In this framework, the goal is to transfer the teacher’s reasoning ability to a smaller student model $\sigma$, while maintaining faithfulness to ground-truth medical knowledge.

Formally, let $\mathcal{D} = \{(x_i, y_i)\}_{i=1}^N$ denote a supervised dataset of medical reasoning problems, where $x_i$ represents the input (e.g., a clinical vignette or question prompt) and $y_i \in \mathcal{Y}$ is the correct answer. For multiple-choice questions, $\mathcal{Y}$ is a finite set of possible answer options. When prompted with an example from $\mathcal{D}$, the teacher model $\mathcal{T}$ generates a sequence of intermediate reasoning steps $r_i = \left(r_i^{(1)}, \ldots, r_i^{(T_i)}\right)$   culminating in a final answer $\hat{y}_i$.

To ensure that the teacher’s generated samples are aligned with the ground truth, we apply \textbf{rejection sampling} guided by a scoring function $S(x_i, r_i, y_i, \hat{y}_i)$. This function evaluates whether the teacher’s final answer is correct, typically defined as:
\begin{equation}
S(x_i, r_i, y_i, \hat{y}_i) =
\begin{cases}
1, & \text{if } \hat{y}_i = y_i, \\
0, & \text{otherwise.}
\end{cases}
\end{equation}

Note that since nearly all of our training tasks are verifiable multiple-choice, our scoring function only needs to compare the ground truth answer with the predicted one. For open-ended tasks, the scoring function may use additional information from the reasoning trace $r_i$ to obtain a score estimate. We then define the \textit{accepted set} of reasoning traces as:
\begin{equation}
\mathcal{R}^+ = \{ (x_i, y_i, r_i) \mid S(x_i, r_i, y_i, \hat{y}_i) = 1 \}.
\end{equation}

These accepted samples represent high-quality teacher rationales that lead to correct final predictions. The student model $\sigma$ is fine-tuned on this distilled dataset $\mathcal{R}^+$ to learn to replicate the teacher’s reasoning trajectories. This ensures that the distilled reasoning traces not only reflect plausible cognitive steps but also reinforce valid conclusions, a critical requirement in medical reasoning tasks.

However, several open design questions remain when performing such distillation in the medical domain, such as which datasets $\mathcal{D}$ to use as question sources and the choice of teacher model $\mathcal{T}$.  We aim to thoroughly explore these questions through the ablation studies described below.

\section{Experimental Setup}

\paragraph{Data Preprocessing} To avoid data contamination with evaluation splits, we performed 16-gram deduplication between text-only benchmarks and all of the questions in our data mixture following the procedure used in S1 \cite{muennighoff2025s1}. To ensure no duplicate images, we hashed all train and test images by hashing the byte data with \texttt{hashlib} and removed any exact overlaps. All image data was preprocessed to have a maximum resolution of 262,144 pixels (512 x 512) using the smart resize algorithm widely used by recent image preprocessors \citep{wang2025internvl3, bai2025qwen2, hong2025glm}. For classification tasks, we performed stratified sampling to balance classes. 

\paragraph{Training Setup} For all data recipe experiments, we fully fine-tuned with batch size 128, learning rate 5e-5, and cosine scheduler with 0.01 linear warmup ratio. We selected \texttt{Qwen2.5-VL-7B-Instruct} as our student model.
We verified the effectiveness of other model families in Sec. \ref{sec:results}.

\section{Data Recipe Experiments}

\begin{figure}
    \centering
    \includegraphics[width=0.95\linewidth]{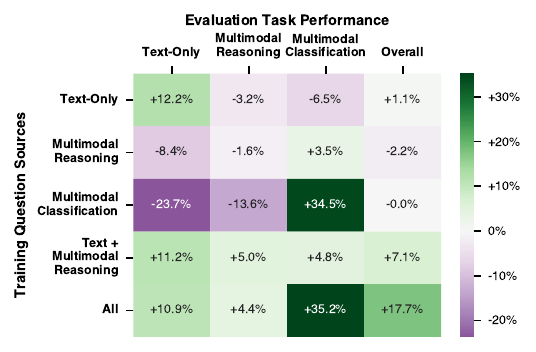}
    \caption{Average performance improvement across downstream task types when training on different question sources. Models perform best when trained on data that matches the downstream task type. Combining sources yields higher and more consistent improvements, suggesting that diverse data sources provide complementary knowledge that enhances generalization.}
    \label{fig:question_sourcing}
\end{figure}

\paragraph{Question Sourcing}
 When performing SFT distillation, the source of knowledge plays a crucial role in determining downstream performance on medical tasks. To investigate this, we grouped our benchmarks and training data into three knowledge-source categories (detailed breakdown in Supplementary Section B): 
\begin{itemize}
    \item \textbf{Text-Only:} Medical reasoning and knowledge questions from USMLE style benchmarks such as MedQA, HeadQA, and MedMCQA.
    \item \textbf{Multimodal Reasoning:} Reasoning-intensive questions about medical imaging from benchmarks such as MMMU-PRO, MedXpertQA, etc. Due to the lack of training splits for multimodal reasoning benchmarks, we used distilled reasoning traces from a subset of the PMC-VQA train split as this question source.
    \item \textbf{Multimodal Classification:} Diagnostic image classification questions from a variety of medical modalities such as Fundus (Aptos, MESSIDOR2), Pathology (BCSS), and MRI (Brain Tumor).
\end{itemize}
We trained student models using different combinations of these knowledge sources and observed the performance on various tasks (Figure \ref{fig:question_sourcing}). We aimed to (1) test cross-source generalization when training on a single dataset, and (2) evaluate whether mixing knowledge sources causes interference and performance degradation. The results show that incorporating training data from a given knowledge source is essential for achieving strong test-time performance within the same category and generalization to unseen sources remains challenging. Moreover, when trained on data drawn from multiple knowledge sources, the student model successfully integrates information from each source without any degradation in overall performance.

\takeawaybox{
Text-only questions are the strongest individual question source. Combining sources boosts generalization without affecting in-domain performance.
}

\begin{table}[ht]
\centering
\resizebox{0.45\textwidth}{!}{%
\begin{tabular}{llcc}
\toprule
\textbf{Category} & \textbf{Task} & \textbf{Direct} & \textbf{CoT} \\
\midrule
\multirow{6}{*}
{\shortstack{Multimodal\\Classification}} 
 & Brain Tumor \cite{bhuvaji2020brain} & 71.32 & 70.56 \\
 & CoronaHack \cite{nasir2023multi} & 62.82 & 66.19 \\
 & Aptos \cite{aptos2019} & 75.98 & 73.60 \\
 & MESSIDOR2 \cite{decenciere2014feedback} & 66.29 & 53.71 \\
 & BCSS \cite{amgad2019structured} & 50.90 & 52.58 \\
 \rowcolor{gray!50} \cellcolor{white} & \textbf{Overall} & \textbf{65.46} & 63.33 \\
\midrule
\multirow{5}{*}{\shortstack{Multimodal\\Reasoning}} 
 & MMMU-PRO (H) \cite{yue2024mmmu} & 12.59 & 31.82 \\
 & NEJM \cite{nejm_image_challenge} & 23.23 & 45.62 \\
 & PMC-VQA \cite{zhang2023pmc} & 33.85 & 49.25 \\
 & MedXpertQA \cite{zuo2025medxpertqa} & 22.65 & 25.90 \\
 \rowcolor{gray!50} \cellcolor{white} & \multicolumn{1}{l}{\textbf{Overall}} & 23.08 & \textbf{38.15} \\
\midrule
\multirow{6}{*}{Text-Only} 
 & MedQA \cite{jin2021disease} & 32.13 & 69.68 \\
 & HeadQA \cite{vilares-gomez-rodriguez-2019-head} & 47.70 & 72.61 \\
 & MedMCQA \cite{pal2022medmcqa} & 34.47 & 53.38 \\
 & MedXpertQA \cite{zuo2025medxpertqa} & 11.43 & 14.04 \\
 & MMLU-PRO (H) \cite{wang2024mmlu} & 20.42 & 51.22 \\
 \rowcolor{gray!50} \cellcolor{white} & \textbf{Overall} & 29.23 & \textbf{52.19} \\
\bottomrule
\end{tabular}}
\caption{Comparison of fine-tuning using different prompting strategies across medical tasks. Chain-of-(CoT) prompting enhances reasoning performance, whereas direct prompting achieves slightly higher accuracy on classification tasks.}
\label{tab:question_prompt}
\end{table}

\paragraph{Question Formatting} Model performance can vary significantly based on the choice of prompt template. We analyzed two common prompting strategies: Chain-of-Thought (CoT) prompting \cite{wei2022chain}, where the model generates an internal reasoning trace before providing its answer, and direct prompting, in which the model responds to the question immediately without a reasoning trace. We hypothesized that CoT prompting would benefit multi-step, reasoning-intensive tasks such as MedQA, while direct prompting might be more suitable for simpler perception tasks, such as diabetic retinopathy grading. To test this, we trained two student models on the same 100k subset of our collected data, varying only the prompt style. As shown in Table \ref{tab:question_prompt}, CoT prompting consistently improves performance on reasoning-heavy tasks (38.15 vs 23.08 on multimodal reasoning), while direct prompting yields better overall results on simpler classification tasks (65.46 vs 63.33). Based on these findings, we adopted CoT prompting for the SFT stage due to its broader applicability and enhanced interpretability. Future work may investigate hybrid approaches to train a model capable of both thinking modes or treat direct prompting as a limiting case of CoT prompting with an empty reasoning trace.

\takeawaybox{
CoT prompting provides significant improvement over direct prompting on reasoning tasks. However, direct prompting outperforms CoT prompting on perceptual tasks such as multimodal classification.
}

\begin{figure}
    \centering
    \includegraphics[width=\linewidth]{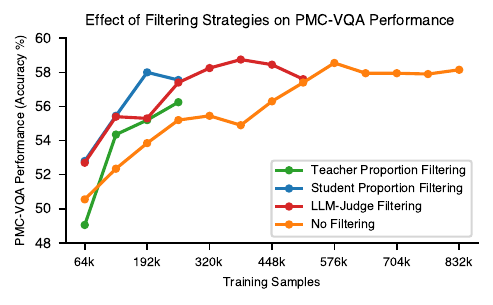}
    \caption{Effect of question filtering on PMC-VQA performance. All filtering strategies improve sample efficiency compared to the no-filtering baseline but have similar peak performance. }
    \label{fig:question_filtering}
\end{figure}

\paragraph{Question Filtering}
Selecting high-quality questions is crucial for improving sample efficiency and filtering out questions that are ambiguous, unanswerable, or contain misleading information. We considered three question filtering strategies designed to balance question difficulty and informativeness:

\begin{itemize}
\item \textbf{Student Model Proportion Filtering: } We prompted the student model with 16 samples per question and discarded those with fewer than 2 or more than 14 correct responses, corresponding to overly difficult or easy questions.
\item \textbf{Teacher Model Proportion Filtering: } We followed the student model proportion filtering strategy but used the teacher model to sample responses, providing a potentially different difficulty estimate.
\item \textbf{LLM-Judge Difficulty Assessment: } We queried \texttt{GPT-4.1-mini} with each question–answer pair and asked it to assign a difficulty rating from 1–10. Based on the resulting distribution, we kept questions within the 3–6 range to focus on moderately challenging examples.
\end{itemize}
Figure~\ref{fig:question_filtering} presents the performance impact of these filtering strategies on PMC-VQA test results. All filtering methods yielded a clear improvement in early training stages, indicating better sample efficiency. However, the unfiltered baseline ultimately achieved a comparable peak performance to the LLM-Judge filtering method. We therefore trained without filtering to maximize data coverage, relying on rejection sampling to downweight low-quality questions. However, question filtering remains promising for RL post-training which benefits from sample efficiency.

\takeawaybox{
Question filtering improves early-training sample efficiency, with student filtering methods performing best. However, all filtering methods converge to similar final performance as the no-filtering baseline.
}

\begin{figure}[!ht]
    \centering
    \includegraphics[width=\linewidth]{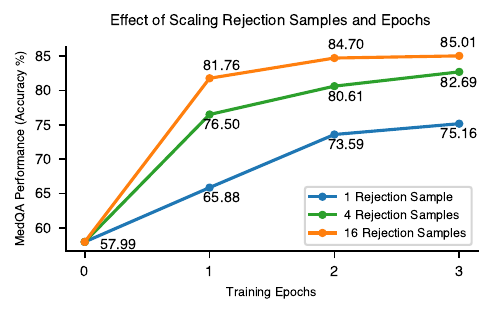}
    \caption{\small Effect of scaling rejection samples and training epochs on MedQA test set performance. Early improvements from additional rejection samples mirror the gains from training for more epochs. However, increasing the number of rejection samples per question consistently raises peak performance, with 16 samples achieving the highest final accuracy.}
    \label{fig:question_samples}
\end{figure}

\paragraph{Question Samples} Prior works such as OpenThoughts \cite{guha2025openthoughts} and II-Medical \cite{ii_medical_2025} have explored using multiple reasoning traces per question, highlighting it as a natural way to expand training data when examples are limited. We conducted a similar study in the medical domain to evaluate the impact of multiple rejection samples per question. Using the training split of the MedQA task, we generated 16 model responses per question and created three experimental settings by limiting the number of valid reasoning traces retained per question to 1, 4, or 16. For each setting, we trained the student model for 3 epochs, resulting in three distinct models and a total of 9 checkpoints. We then evaluated all checkpoints on the unseen MedQA test split to examine performance  (Figure \ref{fig:question_samples}). The results show that in the early epochs, adding more rejection samples per question has an effect similar to training for additional epochs. For example, keeping 1 rejection sample and training for 3 epochs (75.16) has similar performance to keeping 4 rejection samples and training for 1 epoch (76.50). Importantly, peak performance improves by nearly 10\% (from 75.16 to 85.01) when increasing rejection samples per question from 1 to 16, suggesting that diverse reasoning traces serve as a form of regularization that enhances generalization and downstream performance. Based on these findings, we prompted the teacher model 16 times per question and trained for 3 epochs to leverage the benefit of diverse reasoning traces and achieve peak performance.

\takeawaybox{
Given limited, but high quality data, incorporating multiple reasoning traces per question strongly improves robustness and downstream performance.
}

\begin{figure}
    \centering
    \includegraphics[width=\linewidth]{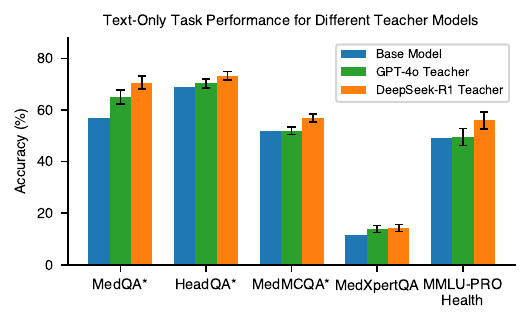}
    \caption{Improvement across text-only medical benchmarks for models distilled from GPT-4o and DeepSeek-R1 teachers. Knowledge is distilled from training splits of tasks marked with an asterisk. Error bars were computed via bootstrapping with 10,000 independent random samples. Both teacher families yield consistent gains over the base model, with DeepSeek-R1 showing larger improvements across tasks, suggesting that reasoning-oriented model families serve as stronger teachers for medical knowledge.}
    \label{fig:teacher_model}
\end{figure}

\paragraph{Teacher Model}
A key decision in our distillation process was the choice of teacher model to generate question-answer pairs. We considered two high-performing models with complementary strengths: GPT-4o \cite{hurst2024gpt} and DeepSeek-R1 \cite{guo2025deepseek}. DeepSeek-R1 demonstrates strong reasoning ability and produces detailed explanations, while GPT-4o offers concise outputs and supports multimodal inputs. Since DeepSeek-R1 is limited to text, we adopted GPT-4o as the teacher for all multimodal data due to its robustness and cost-effectiveness. However, to better understand their relative advantages in the text-only setting, we trained student models on reasoning traces from several text-based reasoning medical tasks, using examples generated by either GPT-4o or DeepSeek-R1. For efficiency, we trained each student for one epoch on a subset of approximately 30k examples, and evaluated them on the corresponding test splits as well as on out-of-distribution text-only benchmarks (results shown in Figure~\ref{fig:teacher_model}). Despite our limited training dataset size, our results show that both teacher models improve text-only task performance, with the DeepSeek-R1 teacher consistently yielding the largest gains across in-domain and out-of-distribution evaluations. GPT-4o represents a multimodal instruction-following model, whereas DeepSeek-R1 is a reasoning-oriented model, which appears better suited for complex medical decision-making tasks.

\takeawaybox{
Both reasoning and instruction-following models are effective teachers. However, reasoning models consistently outperform instruction-following models for medical knowledge distillation.
}

\begin{table*}[!ht]
\centering
\adjustbox{max width=\textwidth}{%
\begin{tabular}{l|l||cccccccc|cccc}
\textbf{Category} & \textbf{Benchmark} &
\rotatebox{45}{\cellcolor{green!20}Qwen2.5} &
\rotatebox{45}{\cellcolor{green!20}HuatuoGPT} &
\rotatebox{45}{\cellcolor{green!20}\shortstack{MedVL\\Thinker}} &
\rotatebox{45}{\cellcolor{green!20}InternVL3.5} &
\rotatebox{45}{\cellcolor{green!20}GLM-4.1V} &
\rotatebox{45}{\cellcolor{green!20}QoQ-Med} &
\rotatebox{45}{\cellcolor{green!20}LingShu} &
\rotatebox{45}{\cellcolor{green!20}\textbf{OctoMed}} &
\rotatebox{45}{\cellcolor{cyan!20}MedGemma} &
\rotatebox{45}{\cellcolor{cyan!20}GPT-4o} &

\rotatebox{45}{\cellcolor{cyan!20}DeepSeek-R1} \\
\midrule
\multirow{3}{*}{Information} & Base LLM &
\raisebox{-0.4\height}{\includegraphics[width=2em]{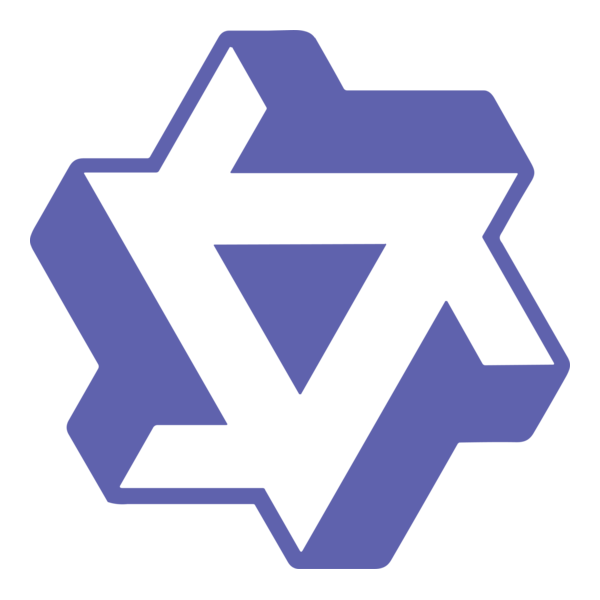}} &
\raisebox{-0.4\height}{\includegraphics[width=2em]{icons/qwen.png}} &
\raisebox{-0.4\height}{\includegraphics[width=2em]{icons/qwen.png}} &
\raisebox{-0.4\height}{\includegraphics[width=2em]{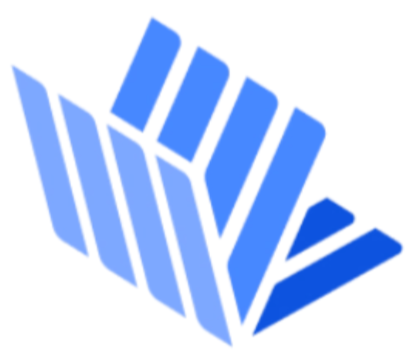}} &
\raisebox{-0.4\height}{\includegraphics[width=2em]{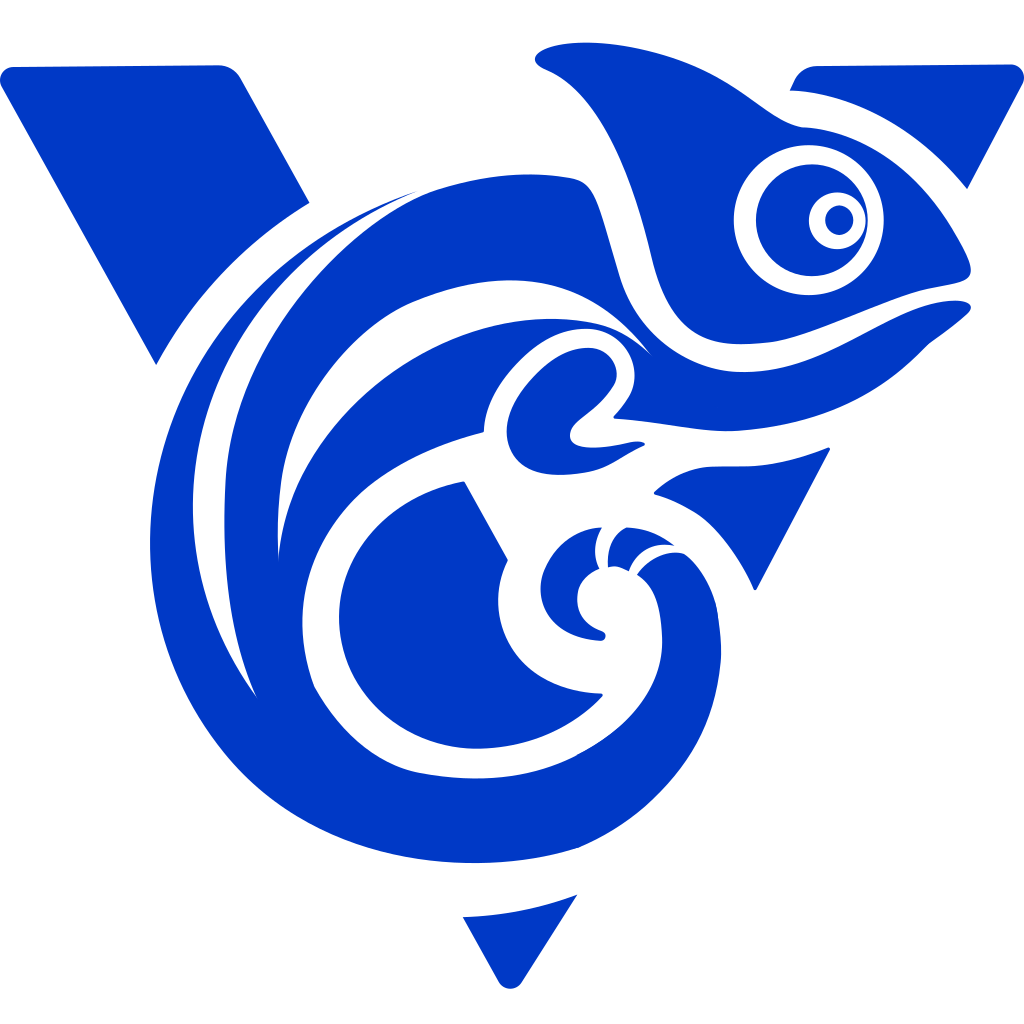}} &
\raisebox{-0.4\height}{\includegraphics[width=2em]{icons/qwen.png}} &
\raisebox{-0.4\height}{\includegraphics[width=2em]{icons/qwen.png}} &
\raisebox{-0.4\height}{\includegraphics[width=2em]{icons/qwen.png}} &
\raisebox{-0.4\height}{\includegraphics[width=2em]{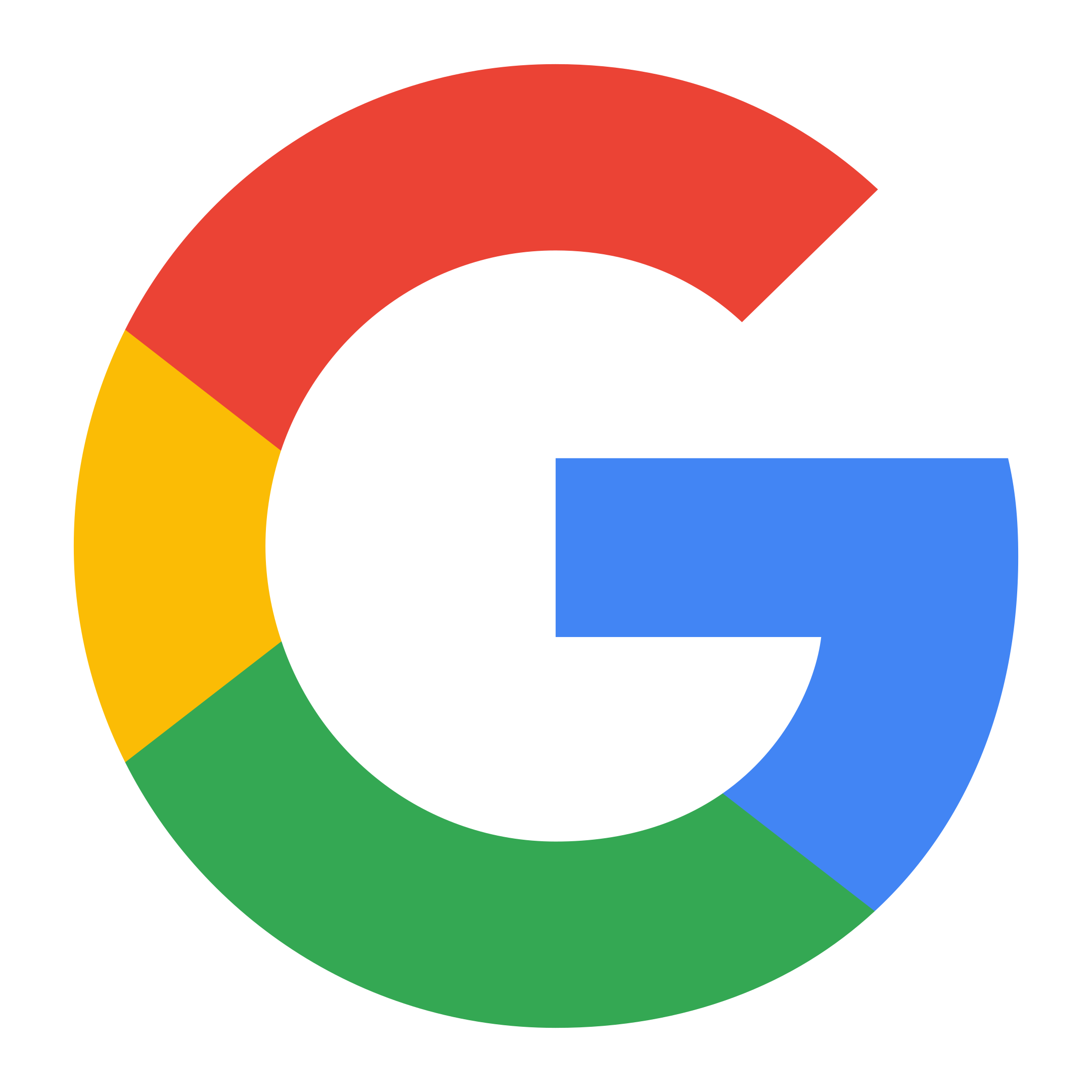}} &
\raisebox{-0.4\height}{\includegraphics[width=2em]{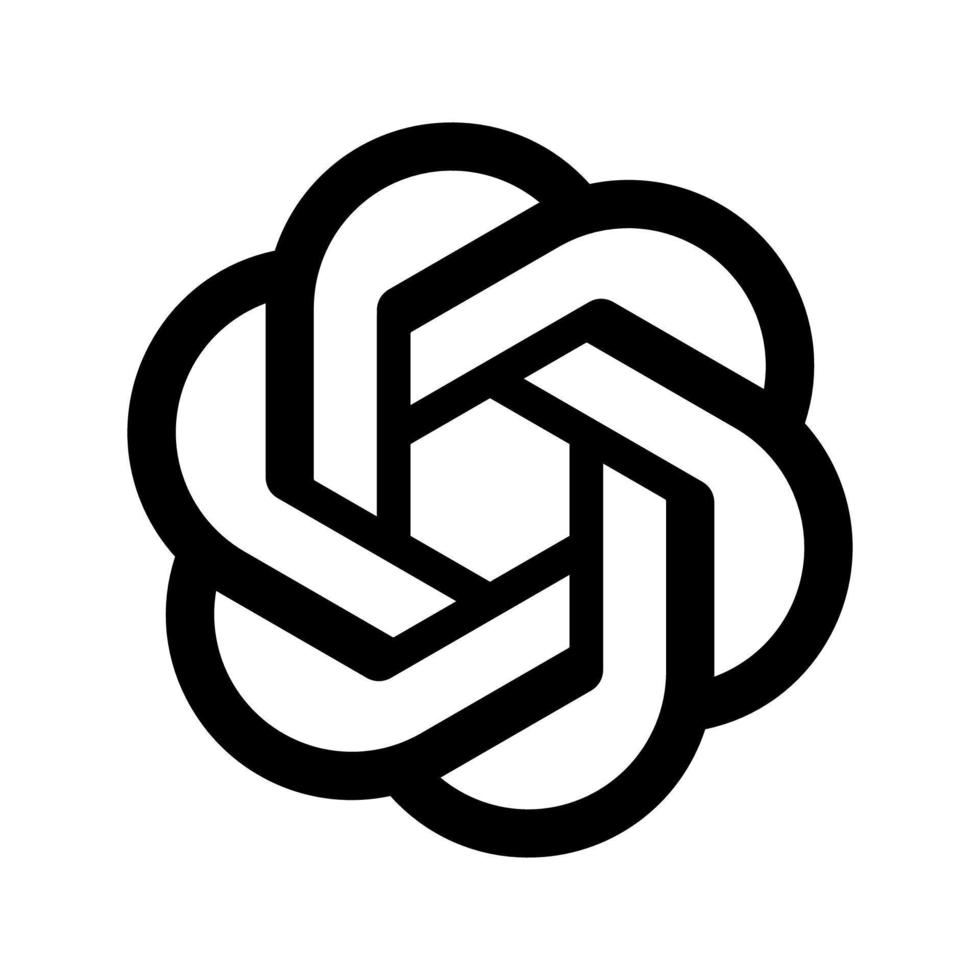}} & 
\raisebox{-0.4\height}{\includegraphics[width=2em]{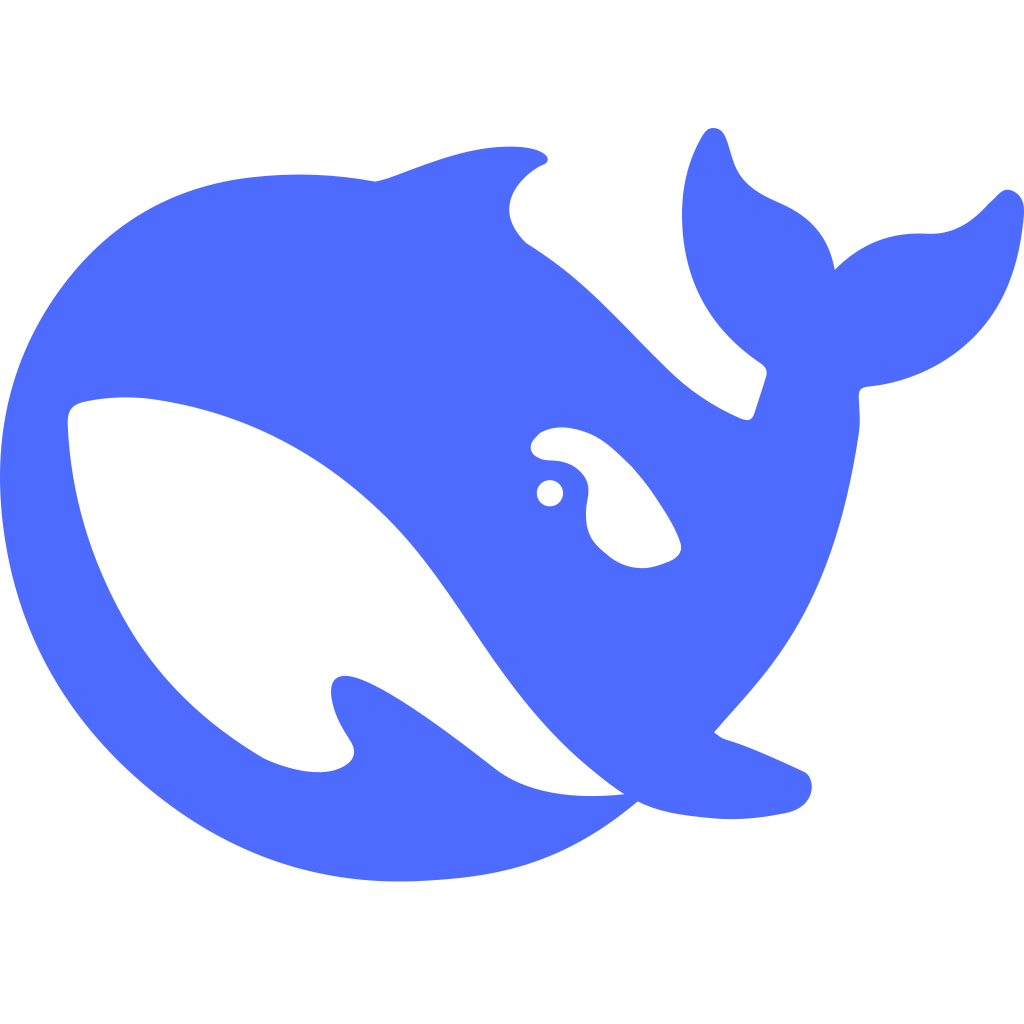}} \\
& Method(s) & SFT & SFT & SFT + RL & SFT + RL & SFT + RL & RL & SFT + RL & SFT & SFT + RL & - & SFT + RL \\
  
& Model Size & 7B & 7B & 7B & 8B & 9B & 7B & 7B & 7B & 27B & $>$ 100B & 671B \\

\midrule

\multirow{6}{*}{Text-Only} 
& MedQA & 57.99 & 55.40 & 58.56 & 73.37 & 76.83 & 58.30 & 62.09 & \textbf{90.81}$^\dagger$ & 85.17 &  90.72 & 93.16 \\
& HeadQA & 69.61 & 67.14 & 71.00 & 81.00 & 81.73 & 70.94 & 69.23 & \textbf{82.36} & 84.56 &  89.56 & 90.93 \\
& MedMCQA & 51.45 & 50.74 & 54.87 & 60.72 & 64.64 & 53.11 & 53.16 & \textbf{72.70} & 70.34 &  77.21 &  79.36 \\
& MedXpertQA & 12.16 & 10.85 & 12.91 & 12.82 & 17.67 & 11.72 & 12.54 & \textbf{24.51} & 21.89 &  30.31 &  37.30 \\
& MMLU-PRO (H) & 49.32 & 44.77 & 52.13 & 62.96 & 68.33 & 49.93 & 50.73 & \textbf{68.75} & 70.83 & 76.06 &  79.39 \\

\rowcolor{gray!50} \cellcolor{white}  &  \textbf{Overall} & 48.10 & 45.78 & 49.89 & 58.17 & 61.84 & 48.80 & 49.55 & \textbf{67.83} & 66.56 &  72.77 & 76.05 \\
\midrule

\multirow{5}{*}{\shortstack{Multimodal\\Reasoning}} 
& MMMU-PRO (H) & 32.87 & 30.00 & 38.95 & 47.90 & \textbf{49.65} & 33.29 & 36.78 & 42.52 &  40.77 & 57.76 &  - \\
& NEJM & 43.76 & 44.06 & 44.31 & 52.90 & 54.81 & 44.01 & 48.43 & \textbf{61.14} &  57.06 & 69.99 &  - \\
& PMC-VQA & 49.66 & 51.46 & 49.88 & 59.55 & 57.05 & 49.79 & 58.76 & \textbf{61.13} & 46.08 &  60.14 &   - \\
& MedXpertQA & 22.47 & 22.63 & 24.81 & 26.10 & 28.65 & 22.74 & 26.19 & \textbf{36.65} &  33.13 & 44.38 &  - \\
\rowcolor{gray!50} \cellcolor{white} & \textbf{Overall} & 37.19 & 37.04 & 39.49 & 46.61 & 47.54 & 37.46 & 42.54 & \textbf{50.36} &  44.25 & 58.07 &  - \\
\midrule

\multirow{6}{*}{\shortstack{Multimodal\\Classification}} 
& Brain Tumor & 27.82 & 37.16 & 31.07 & 64.47 & 47.72 & 55.33 & 78.99 & \textbf{80.86} & 58.33 &  65.99 &  - \\
& CoronaHack & 34.17 & 42.76 & 36.41 & 38.62 & 39.90 & 50.13 & 42.24 & \textbf{71.22} & 58.27 &  41.70 &  - \\
& Aptos & 26.71 & 40.95 & 25.08 & 51.96 & 42.76 & 64.60 & 56.83 & \textbf{75.43} & 49.20 &  66.58 & - \\
& MESSIDOR2 & 20.63 & 33.66 & 22.40 & 27.71 & 26.86 & \textbf{59.14} & 42.86 & 55.20 & 53.09 &  45.26 &  - \\
& BCSS & 43.12 & 43.72 & 48.01 & 50.95 & 52.84 & 52.02 & 52.40 & \textbf{53.72} & 35.95 &  50.28 &  - \\
\rowcolor{gray!50} \cellcolor{white} & \textbf{Overall} & 30.49 & 46.39 & 32.59 & 46.39 & 41.76 & 56.24 & 54.66 & \textbf{67.29} & 50.97 & 53.96 &  - \\

\bottomrule
\end{tabular}
}

\caption{Performances across Text-only, Multimodal Reasoning, and Multimodal Classification medical benchmarks. Models with green background are OSS smaller models ($<$10B parameters), and models with cyan background are large proprietary models. For the OSS smaller models, \raisebox{-0.3\baselineskip}{\includegraphics[width=1.5em]{icons/qwen.png}} refers to Qwen2.5-VL-7B-Instruct, \raisebox{-0.3\baselineskip}{\includegraphics[width=1.5em]{icons/internvl.png}} refers to InternVL3.5-8B, and \raisebox{-0.3\baselineskip}{\includegraphics[width=1.5em]{icons/glm.png}} refers to GLM-4.1V-Thinking-9B. \\ $^\dagger$ 10-sample majority vote ensemble result.}
\label{tab:performances}

\end{table*}

\section{Results}
\label{sec:results}

Our final model was fully fine-tuned on our dataset of 8 million reasoning traces for 3 epochs. We used \texttt{llamafactory} \cite{zheng2024llamafactory} as our training framework, with a learning rate of 5e-5, effective batch size of 512 with cosine learning rate scheduler with linear warmup ratio of 0.1. We used \texttt{Qwen2.5-VL-7B-Instruct} as our student model.

\begin{figure}[!ht]
    \centering
    \includegraphics[width=0.9\linewidth]{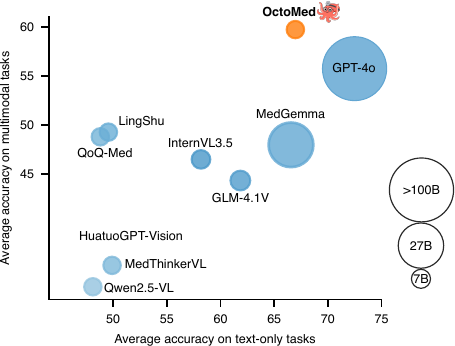}
    \caption{Performance of strong multimodal reasoning models on text-only and multimodal medical benchmarks. Size of each circle is proportional to the number of model parameters.}
    \label{fig:model_performance_scatter}
\end{figure}

\paragraph{Evaluation Setup}
We evaluated all models in our codebase by computing an average accuracy score across 5 independent runs with different sampling seed. We used a temperature of 0.6 and top-p of 0.95, and followed the suggested prompt templates reported by each model for its own evaluation. We used \texttt{vllm} \cite{kwon2023efficient} as our inference engine, allowing a max response length of 8192 tokens (plus additional tokens corresponding to 10 multimodal image inputs of max resolution 262,144 pixels). For reasoning models with a think-first response format, we also enabled a forced-exiting mechanism described in \cite{muennighoff2025s1} if the model failed to terminate its internal reasoning trace. In this case, we appended an end of think token to the unfinished reasoning trace, encouraging the model to answer the question given its thoughts so far.

\begin{figure*}[!ht]
    \centering
    \includegraphics[width=\linewidth]{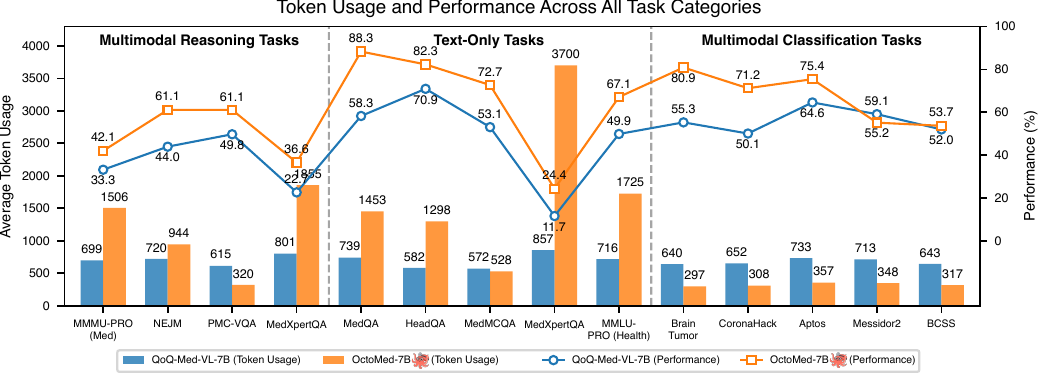}
    \caption{\small Average response length (in tokens) across in-domain and out-of-distribution (OOD) benchmarks. Compared to QoQ-Med-VL-7B, \ours produces longer reasoning traces on challenging reasoning tasks such as MedXpertQA and MMMU-PRO, and shorter reasoning traces on simpler tasks such as PMC-VQA and Brain Tumor. This behavior suggests that \ours adapts its reasoning length to task complexity, whereas QoQ-Med exhibits relatively uniform token usage across tasks. }
    \label{fig:token_usage}
\end{figure*}

\paragraph{Overall Performance}
After training on 8 million structured reasoning traces, our model achieves state-of-the-art performance on various benchmarks and demonstrates strong generalization to unseen tasks. As shown in Figure \ref{fig:model_performance_scatter} and Table \ref{tab:performances}, \ours greatly exceeds the performance of similar size open-source models on all benchmark categories. For models several times larger, \ours remains competitive, even outperforming MedGemma-27B overall on every benchmark category despite being 4x smaller. In multimodal medical classification tasks specifically, \ours surpasses GPT-4o which was used as a teacher model (67.29 vs 53.96). These results indicate that scaling high-quality reasoning traces can rival or exceed the performance of much larger domain-specialized models, providing an efficient path toward robust multimodal and text-only medical generalization.

\paragraph{Emergence of Task-Aware Thinking}
Since our training data mixture was curated using both reasoning (DeepSeek-R1) and instruction-following (GPT-4o) teacher models, it spans a wide range of reasoning trace lengths. To investigate how this design choice impacts response length on out-of-distribution (OOD) tasks, we analyzed the average number of output tokens produced by \ours across various in-domain and OOD benchmarks, shown in Figure \ref{fig:token_usage}, and compared it with a strong baseline trained from the same base model. Notably, even without exposure to multimodal reasoning tasks such as MedXpertQA, \ours generates substantially longer reasoning traces on these benchmarks, a behavior which appears to correlate with task difficulty. For example, \ours spends an average of just 320 tokens reasoning on PMC-VQA, consistent with the fact that many of its questions are relatively straightforward and do not require multi-step logical inference. In contrast, QoQ-Med maintains similar reasoning lengths across tasks, showing less sensitivity to task complexity. This dynamic reasoning length opens a potential way to use \ours in post-training pipelines to measure task difficulty and filter questions by analyzing its reasoning trajectory lengths.

\paragraph{Comparison to Existing Datasets}
To evaluate the effectiveness of our SFT dataset, we compared it against three existing multimodal medical SFT datasets. As shown in Figure~\ref{fig:data_dist}, we fixed the student model to \texttt{Qwen2.5-VL-7B-Instruct} and followed the fine-tuning configurations specified in the original works, tracking performance on evaluation tasks throughout training. For checkpoints trained with the PubMedVision and LLaVA-Med datasets, we employed \texttt{GPT-4.1-mini} as an LLM judge, as these models struggled to follow formatting instructions during evaluation. The results demonstrate that even within the first 16\% of the training schedule, \ours achieves substantial performance improvements. In contrast, fine-tuning on other datasets yielded only marginal gains and failed to surpass the baseline performance of the original student model in multimodal and text-only reasoning.
\begin{figure*}[!t]
    \centering
    \includegraphics[width=\linewidth]{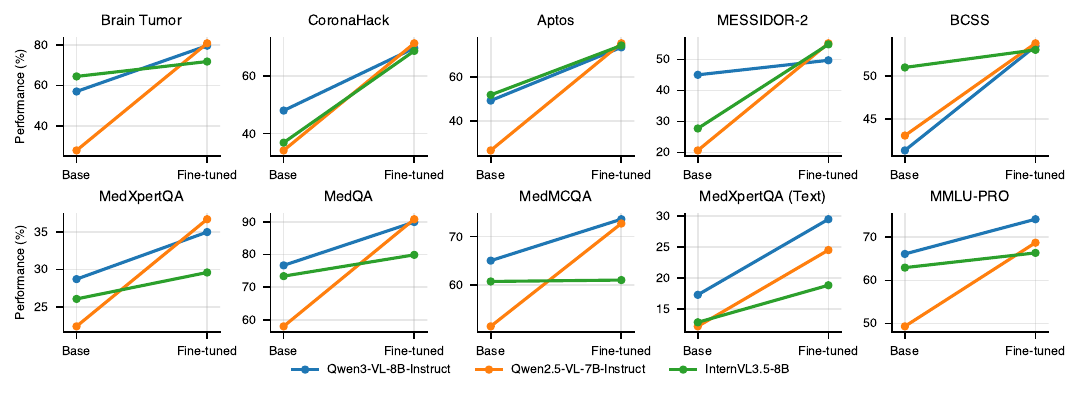}
    \caption{Result of finetuning separate model families on our dataset across several representative tasks. The top row contains multimodal classification tasks, and the bottom row shows various reasoning tasks. Performance gains are consistent across model families for classification tasks, but for reasoning tasks, instruction-following models benefit more from SFT than those already post-trained for reasoning.}
    \label{fig:scaling_model_axis}
\end{figure*}

\paragraph{Model Families}
Due to recent investigations on the possibility of data leakage in the Qwen2.5 series \cite{wu2025reasoning, shao2025spurious}, we performed SFT on our dataset with different model families to verify its effectiveness. In addition to \texttt{Qwen2.5-VL-7B-Instruct}, we selected two more strong vision-language models, one post-trained for reasoning and the other for instruction-following: \texttt{InternVL3.5-8B} and \texttt{Qwen3-VL-8B-Instruct} respectively. As shown in Figure \ref{fig:scaling_model_axis}, we observed similar performance gains across model families, with some variations depending on the qualities of the base model. Finetuning base models which had already been post-trained for reasoning (e.g. InternVL3.5-8B) resulted in reduced text-only and multimodal-reasoning gains compared to finetuning instruction-following models such as Qwen3-VL-8B-Instruct. This finding suggests that SFT is more effective when applied prior to reinforcement learning.

\paragraph{Qualitative Analysis}
We qualitatively examine reasoning traces generated by our model across diverse task types and difficulty levels (Supplementary Section~A). The examples highlight the following key properties:
\begin{itemize}
\item \textbf{Task Versatility:} \ours extends beyond multiple-choice VQA, effectively handling open-ended and descriptive reasoning tasks (Figure~S1).
\item \textbf{Modality Versatility:} \ours generalizes across text-only tasks and multiple imaging modalities, such as MRI, fundus, dermatology, and pathology, while maintaining coherent multimodal reasoning (Figure~S2).
\item \textbf{Task-Aware Thinking:} \ours adapts its reasoning depth and structure to the complexity and format of each task, demonstrating contextual awareness and efficient problem solving (Figure~S3).
\end{itemize}
These qualitative findings demonstrate that our SFT recipe produces a versatile model capable of consistent, context-aware reasoning across modalities and task formats.

\section{Related Work}

Recent vision-language models have demonstrated that high-quality instruction-tuning data can elicit strong reasoning abilities through supervised fine-tuning (SFT) alone \cite{yue2023mammoth, huang2025vision, yang2025r1, xu2025llava, guo2025mammoth}. In medicine, LLaVA-Med \cite{li2023llava} and HuatuoGPT-Vision \cite{chen2024huatuogpt} pioneered multimodal reasoning by distilling datasets such as PubMedVision from biomedical image–caption pairs, while II-Medical \cite{ii_medical_2025} further explored this direction with large-scale text-only reasoning traces. Building on these SFT foundations, recent efforts combine SFT with reinforcement learning (RL) for verifiable reasoning \cite{hong2025glm, bai2025qwen2, peng2025skywork, liu2025x}, with medical variants such as MedVLThinker \cite{huang2025medvlthinker}, LingShu \cite{xu2025lingshu}, ReasonMed\cite{sun-etal-2025-reasonmed}, MedGemma \cite{sellergren2025medgemma} scaling across diverse modalities.

In contrast, some models (e.g., MedVLM-R1 \cite{pan2025medvlm}, Med-R1 \cite{lai2025med}, Med-RLVR \cite{zhang2025med}, and QoQ-Med \cite{dai2025qoq}) skip SFT entirely, leveraging RL-based objectives for improved sample efficiency. While prior work has primarily focused on new training objectives or large curated datasets, comparatively little attention has been given to the composition of the training mixture itself. Recent efforts such as Honeybee \cite{bansal2025honeybee} and FineVision \cite{wiedmann2025finevision} have explored data recipes for general vision–language reasoning, whereas our work targets medical multimodal reasoning. We introduce a principled data recipe that balances modality coverage and reasoning difficulty to enable scalable, high-quality multimodal training in the medical domain.
\section{Conclusion}

We present \ours, an exploration of training and data curation strategies for multimodal reasoning in the medical domain. Our preliminary findings suggest that different design choices for the data recipe can significantly impact performance, and scaling up this data recipe achieves state-of-the-art results on downstream tasks. Experiments indicate that supervised fine-tuning (SFT) alone is sufficient to produce high-quality reasoning models in the medical domain. Moreover, the final finetuned model exhibits task-aware reasoning calibration, dynamically adjusting its reasoning strategies based on the task at hand. Although our work focuses on SFT, extending \ours with reinforcement learning represents a promising direction for enhancing robustness across diverse clinical scenarios.
{
    \small
    \bibliographystyle{ieeenat_fullname}
    \bibliography{main}
}

\clearpage
\appendix
\clearpage
\setcounter{page}{1}
\maketitlesupplementary

\section{Case Studies}
We qualitatively analyze the performance of \ours on various multimodal medical queries. In particular, we note the following capabilities:
\renewcommand{\thefigure}{S1-\arabic{figure}}
\setcounter{figure}{0}
\begin{figure*}[h]
    \begin{AIbox}{Medication Assessment Example}

   \noindent

\scriptsize
\begin{alltt}
A 55-year old woman with active smoking history, type II diabetes, hypertension, history of coronary artery
disease was recently hospitalized for heart failure with an ejection fraction of 35\%. She is presenting for
follow-up post discharge. Her discharge medications included lisinopril 10 mg daily, carvedilol 2.5 mg twice
daily, and eplerenone 25 mg daily, in addition to her usual medications for diabetes. In clinic, she reports
feeling symptomatic with dyspnea after going up 1 flight of stairs though is asymptomatic at rest.
Vital signs at rest: BP 130/90, HR 75, RR 10, Temperature 97.2F, SpO2 98\% on room air. Labs are remarkable
for Na 132 mEq/L, K 4.9 mEq/L, BUN 22 mg/dL, Cr 1.04 mg/dL.
The patient reports having a friend on Entresto. Please review whether Entresto is indicated in this patient and
why or why not.

\end{alltt}

 \footnotesize \textbf{\ours:} \scriptsize \textbf{\texttt{<think>}} Okay, let's tackle this case step by step. So the patient is a 55-year-old woman with multiple comorbidities: active smoking, type II diabetes, hypertension, history of coronary artery disease, and now she's been hospitalized for heart failure with reduced ejection fraction (HFrEF) at 35\%. She's on several meds including lisinopril, carvedilol, eplerenone, and others for diabetes. \\

First, I need to recall the guidelines for managing HFrEF. The mainstay therapies include ACE inhibitors (or ARBs if intolerant), beta-blockers, and mineralocorticoid receptor antagonists (MRAs) like eplerenone. Additionally, SGLT2 inhibitors have become important lately. But wait, the question is about Entresto, which is sacubitril/valsartan. That's an ARNI—angiotensin receptor-neprilysin inhibitor. \\

So, when is Entresto indicated? From what I remember, it's used in HFrEF patients as an alternative to ACE inhibitors or ARBs. The PARADIGM-HF trial showed that sacubitril/valsartan was better than enalapril in reducing cardiovascular death and hospitalization for HF. The indication typically requires NYHA class II-IV symptoms despite being on optimal therapy including an ACE-I or ARB, MRA, and beta-blocker. Also, they need to be stable hemodynamically, no hypotension, renal issues, hyperkalemia, etc. \\

Looking at this patient: She has HFrEF EF 35\%, so definitely reduced. She's symptomatic with dyspnea on exertion (1st flight of stairs). Post-discharge, but maybe still in NYHA class II. Her current meds include lisinopril (ACE-I), carvedilol (beta-blocker), and eplerenone (MRA). So she's already on all three pillars. However, the next step might be substituting the ACE-I with Entresto if appropriate. \\

But there are contraindications. Let's check her labs: Sodium is low at 132, potassium 4.9 (which is okay, upper limit but acceptable). Creatinine is 1.04, BUN 22. So renal function seems okay, though creatinine isn't super high. She doesn't have significant hyperkalemia here. Her blood pressure is controlled at 130/90. No mention of angioedema or cough from lisinopril, so probably can switch. \\

Wait, but the guidelines say that Entresto should replace the ACE-I or ARB, not add to them. Since she's on lisinopril, switching to Entresto would make sense. But does she meet criteria? She has persistent symptoms despite being on those meds. The fact that she's still symptomatic (dyspnea on exertion) suggests that her current regimen isn't fully controlling her symptoms. Therefore, moving to Entresto could provide additional benefit. \\

Another consideration is that she has diabetes. Entresto can be used in diabetics, although monitoring for hyperkalemia and renal function is needed. Her potassium is normal, and her creatinine is slightly elevated but not critically. Her BUN is 22 mg/dL and her creatinine is 1.04 mg/dL. To estimate her GFR using the CKD-EPI formula for a 55-year-old female, we use: eGFR = 141 $\times$ (Scr)$^{-1.209}$ $\times$ (Age)$^{-0.203}$ $\times$ 0.742. Assuming Scr = 1.04 mg/dL and age = 55 years, we approximate: 141 $\times$ (1.04$^{-1.209}$) $\approx$ 141 $\times$ 0.82 $\approx$ 114, and then 114 $\times$ 0.742 $\approx$ 83. Thus, her estimated GFR is approximately 83 mL/min/1.73 m$^2$. This corresponds to stage 1 chronic kidney disease but is considered mild, and therefore Entresto would be safe here unless other issues arise. \\

Also, she's on eplerenone. Combining MRAs with Entresto increases risk of hyperkalemia. Her potassium is 4.9, which is within normal range (normal up to 5.0 or 5.1 depending on lab). So maybe manageable with close monitoring. Guidelines recommend avoiding combination unless necessary, but in some cases, benefits outweigh risks. Given her EF is 35\%, and symptoms persisting, adding Entresto while continuing eplerenone may be considered, but perhaps with caution. \\

Wait, but standard practice is to substitute ACE-I/ARB with Entresto. If she's already on eplerenone and carvedilol, then replacing lisinopril with Entresto would be the way to go. Then continue eplerenone. Alternatively, sometimes MRA can be held temporarily when starting Entresto due to hyperkalemia risk, but recent studies suggest that continuation may be possible with monitoring. The 2022 AHA/ACC/HFSA guidelines state that combining Entresto with MRA is reasonable in certain patients, especially if they're stable, but requires close monitoring of potassium and renal function. \\

In this case, since her potassium is normal and renal function is preserved, adding Entresto while keeping eplerenone might be acceptable. However, some clinicians prefer to hold the MRA for a week after initiating Entresto and monitor. But given that she's symptomatic, the benefit of Entresto likely outweighs the risk. \\

Alternatively, if her potassium were higher, maybe hold eplerenone. But here, K is 4.9, which is just under the cutoff (usually 5.0 or 5.5). So perhaps proceed with Entresto substitution, keep eplerenone, monitor K closely. \\

Other factors: she's an active smoker. Smoking cessation counseling is essential. Also, optimizing diabetes control, encouraging exercise, low-sodium diet, etc. But focusing on medication: Entresto is indicated here because she's on ACE-I, beta-blocker, MRA, and remains symptomatic. Replacing ACE-I with Entresto would align with guideline-directed medical therapy (GDMT) for HFrEF.\textbf{\texttt{</think>}} \\

\textbf{Entresto (sacubitril/valsartan)} is \textbf{indicated} for this patient based on the following rationale: \\

\textbf{Key Indications for Entresto:} \\

\begin{enumerate}
    \item \textbf{HFrEF with Persistent Symptoms:}
The patient has heart failure with reduced ejection fraction (HFrEF; EF 35\%) and remains symptomatic (dyspnea on exertion despite optimal therapy). This aligns with NYHA Class II symptoms, where Entresto improves outcomes compared to ACE inhibitors (e.g., enalapril).
\item \textbf{Guideline-Directed Medical Therapy (GDMT):}
Current ACC/AHA/HFSA guidelines recommend replacing ACE inhibitors or ARBs with Entresto in HFrEF patients who remain symptomatic despite $\geq$3 months of evidence-based therapy (ACE-I $+$ beta-blocker $+$ MRA). This patient is on lisinopril (ACE-I), carvedilol (beta-blocker), and eplerenone (MRA), fulfilling these criteria.
\item \textbf{Superior Efficacy:}
The PARADIGM-HF trial demonstrated Entresto reduces cardiovascular mortality and HF hospitalizations more effectively than enalapril. Substituting lisinopril with Entresto addresses residual symptoms and optimizes prognosis.
\end{enumerate}

\end{AIbox}
    \caption{In addition to multiple-choice VQA, \ours can be prompted to perform other tasks such as medication assessment, shown above.}
    \label{tab:medication_assessment}
\end{figure*}
\begin{figure*}[h]
    \begin{AIbox}{Chest X-Ray Report Generation Example}

   \noindent
\begin{center}
\includegraphics[width=0.7\textwidth]{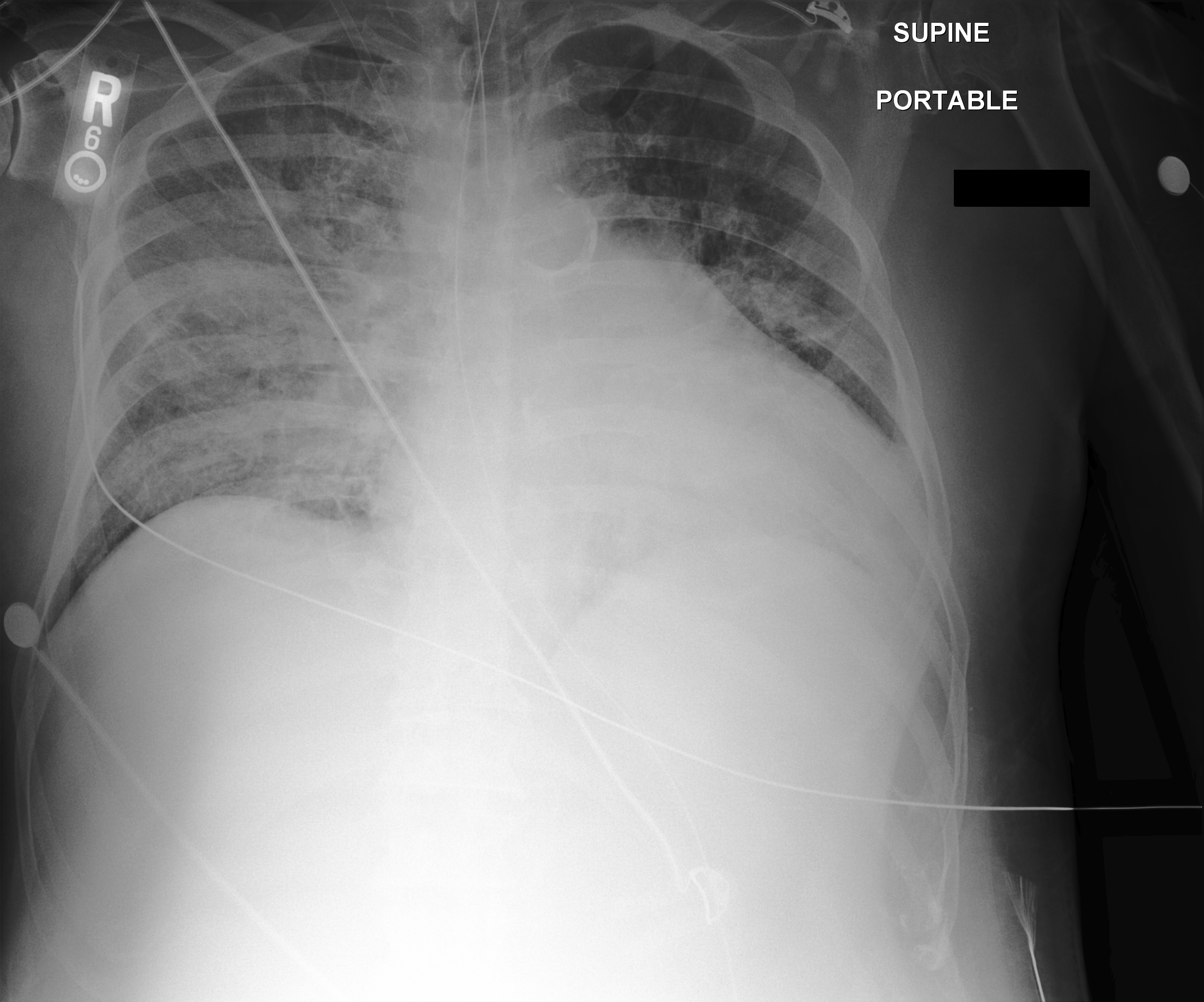}

\end{center}

\begin{center}
    \begin{alltt}
\makebox[\textwidth][c]{Describe the findings in this chest x-ray.}
\end{alltt}
\end{center}

 \footnotesize \textbf{\ours:} \scriptsize \textbf{\texttt{<think>}}When analyzing this chest X-ray, I begin by systematically examining the image to identify any abnormalities or significant findings.

 \begin{enumerate}
     \item \textbf{Image Orientation and Quality:}
     
     The X-ray is labeled as "SUPINE" and "PORTABLE," indicating that it was taken with the patient lying down (supine position) using a portable machine. This positioning can affect the appearance of certain structures, such as the heart size and lung markings, so these factors must be considered during interpretation.
     \item \textbf{Lines and Tubes:}
     
     Multiple lines and tubes are visible:
A central venous catheter (CVC) is seen entering from the right side and traveling toward the mediastinum.
An endotracheal tube is present, likely used for mechanical ventilation.
Additional monitoring electrodes and wires are also visible but do not interfere significantly with the assessment.
     \item \textbf{Cardiac Silhouette:}
     
     The cardiac silhouette appears enlarged. While supine films may exaggerate heart size due to magnification effects, the enlargement here seems disproportionate even accounting for positional changes. This suggests cardiomegaly or possible pericardial effusion.
     \item \textbf{Pulmonary Findings:}
     
     There is increased opacity throughout both lungs, particularly prominent in the lower zones bilaterally. These opacities appear patchy and diffuse, consistent with alveolar filling processes.
The pattern of opacification resembles pulmonary edema, which could result from fluid accumulation in the interstitial and alveolar spaces.
No clear focal consolidation or cavitation is noted, making pneumonia less likely as the primary cause of these findings.
The vascular markings are more pronounced, suggesting vascular congestion, which aligns with pulmonary edema.
     \item \textbf{Pleural Spaces:}
     
     Both costophrenic angles are obscured, raising suspicion for pleural effusions. In the supine position, pleural effusions layer posteriorly and can obscure the diaphragms rather than creating the classic meniscus sign seen on upright films.
     \item \textbf{Diaphragm and Abdomen:}

    The diaphragmatic contours are poorly defined, especially on the left side, further supporting the presence of pleural effusions.
    Gas patterns within the abdomen suggest bowel loops filled with air, which are incidental findings unrelated to the thoracic pathology.
     \item \textbf{Differential Diagnosis:}
     
     The combination of bilateral diffuse opacities, obscured costophrenic angles, and an enlarged cardiac silhouette strongly suggests pulmonary edema secondary to congestive heart failure.
Other considerations include acute respiratory distress syndrome (ARDS), though ARDS typically lacks significant cardiomegaly unless complicated by cardiac dysfunction.
     \item \textbf{Summary:}
     
     Based on the observed features—cardiomegaly, bilateral pulmonary opacities consistent with alveolar flooding, and evidence of pleural effusions—the most likely diagnosis is pulmonary edema. This conclusion fits well with the clinical context of a critically ill patient requiring intensive support (e.g., intubation and CVC).
     
 \end{enumerate}

By carefully evaluating each component of the image step-by-step, I arrive at the final interpretation. 

\textbf{\texttt{</think>}} \\

The chest x-ray shows signs of extensive bilateral pulmonary edema, characterized by widespread opacification of both lung fields, indicative of fluid accumulation within the alveolar spaces. Additionally, there is evidence of pleural effusions, as suggested by blunted costophrenic angles and obscuration of the diaphragmatic borders. Cardiomegaly is apparent, indicated by an enlarged cardiac silhouette. The presence of medical devices like the central venous catheter and endotracheal tube indicates the patient is receiving intensive care.
    \end{AIbox}
    \caption{Example of prompting \ours to perform a Chest X-Ray report generation task.}

    \label{tab:cxr_report_gen}
\end{figure*}

\renewcommand{\thefigure}{S2-\arabic{figure}}
\setcounter{figure}{0}
\begin{figure*}[h]
    \begin{AIbox}{Histopathology Example}

   \noindent
   \begin{center}
\begin{minipage}[t]{0.2\textwidth}
    \includegraphics[width=\linewidth]{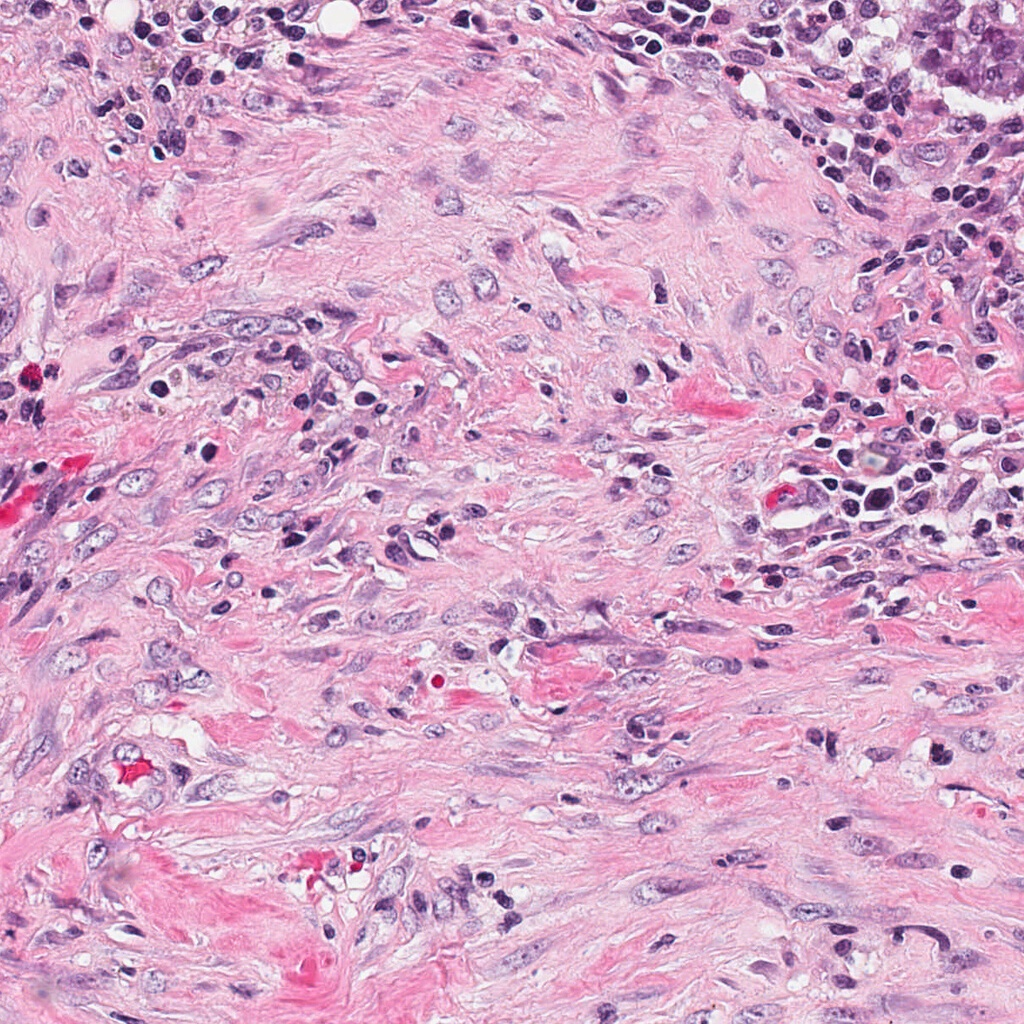}
\end{minipage}%
\hfill
\begin{minipage}[t]{0.75\textwidth}
    \vspace{-3.3cm}
    \footnotesize
    \begin{alltt}
Above is a histopathological image patch from breast cancer tissue.
What type of tissue is shown in this image patch?
A: Tumor
B: Stroma \textcolor{ForestGreen}{(ground truth)}
C: Inflammatory
D: Necrosis

Please reason step-by-step, and put your final answer within \textbackslash \textbackslash boxed\{\}.
    \end{alltt}
    
\end{minipage}
\end{center}

\scriptsize

 \footnotesize \textbf{\ours:} \scriptsize \textbf{\texttt{<think>}} To determine the type of tissue shown in the histopathological image, I need to carefully analyze the visual characteristics. The image appears to be stained with hematoxylin and eosin (H\&E), which is commonly used for histological examination. \\

First, I notice a dense pink background with spindle-shaped cells that are characteristic of stromal tissue. The pink areas represent extracellular matrix components, such as collagen, which are abundant in the stroma. The spindle-shaped cells are fibroblasts or myofibroblasts, which are also typical of stromal tissue. \\

Next, I observe scattered darker purple-stained nuclei. These nuclei are elongated and irregular, which is consistent with stromal cells rather than the round or pleomorphic nuclei typically seen in tumor cells. Tumor tissue would generally exhibit a higher degree of nuclear atypia, mitotic figures, and cellular crowding, which are not prominent here. \\

Inflammatory tissue would show a significant infiltration of immune cells such as lymphocytes, plasma cells, or neutrophils, which would appear as small, round, dark-stained nuclei. While there are some immune cells in the image, they are sparse and not the dominant feature. \\

Necrotic tissue would show areas of cell death with ghost-like cell outlines, loss of nuclear staining, and eosinophilic debris. There is no evidence of necrosis in this image, as the cells appear intact with visible nuclei and organized structure. \\

Based on these observations, the tissue in the image is most consistent with stroma, as it shows abundant extracellular matrix, spindle-shaped fibroblasts, and a lack of features characteristic of tumor, inflammatory, or necrotic tissue. \textbf{\texttt{</think>}} \\
 
\textbf{\textbackslash \textbackslash boxed\{B: Stroma\}}
\end{AIbox}
    \caption{Example of a multimodal classification based on histopathological tissue.}
    \label{tab:bcss_example}
    
\end{figure*}
\begin{figure*}[h]
    \begin{AIbox}{Chest X-Ray Example}

   \noindent
   \begin{center}
\begin{minipage}[t]{0.2\textwidth}
    \includegraphics[width=\linewidth]{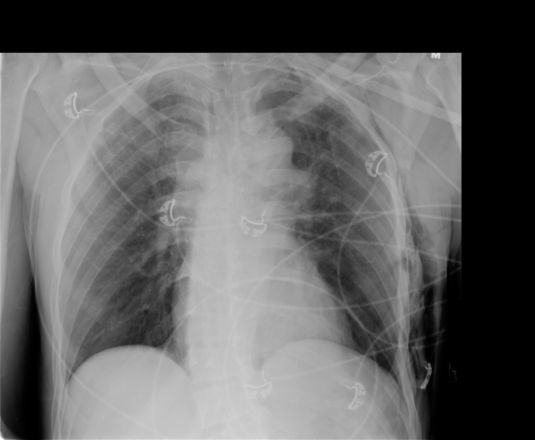}
\end{minipage}%
\hfill
\begin{minipage}[t]{0.75\textwidth}
    \vspace{-2.6cm}
    \footnotesize
    \begin{alltt}
What finding is observed over the left lateral chest on this chest X-ray?
A: Subcutaneous emphysema \textcolor{ForestGreen}{(ground truth)}
B: Pleural effusion
C: Rib fracture
D: Mediastinal mass

Please reason step-by-step, and put your final answer within \textbackslash \textbackslash boxed\{\}.
    \end{alltt}
    
\end{minipage}
\end{center}

\scriptsize

 \footnotesize \textbf{\ours:} \scriptsize \textbf{\texttt{<think>}}To analyze the findings in this chest X-ray, I need to carefully evaluate the image and consider each option systematically.\\

First, let me examine the area of interest: the left lateral chest. The question specifically asks about an abnormality in this region, so my focus will be there.\\

\begin{enumerate}
    \item  \textbf{Option A: Subcutaneous emphysema}
    Subcutaneous emphysema appears as streaky or radiolucent (dark) areas within the soft tissues due to air trapped under the skin. On the X-ray, I should look for unusual dark lines or pockets outside the bony thoracic cage, particularly in the soft tissue planes. In this image, there does appear to be some lucency extending along the left lateral chest wall, which could suggest subcutaneous emphysema. However, I need to rule out other possibilities before concluding.
    \item \textbf{Option B: Pleural effusion}
    Pleural effusion typically manifests as blunting of the costophrenic angle or a meniscus-shaped opacity at the lung base. It would cause increased density (whiteness) rather than lucency. Here, the left lateral chest shows no significant fluid accumulation or opacification suggestive of pleural effusion, so this is less likely.
    \item \textbf{Option C: Rib fracture}
    Rib fractures may show discontinuity or irregularity in the cortical outline of the ribs. While it can sometimes be subtle, I don’t see any clear evidence of rib fractures in the left lateral chest on this X-ray. Additionally, rib fractures alone wouldn't explain the lucency seen here unless associated with secondary complications like pneumothorax or subcutaneous emphysema.
    \item \textbf{Option D: Mediastinal mass}
    A mediastinal mass would present as a well-defined opacity in the central chest, near the heart or great vessels. This finding is not located laterally in the chest, making this option unlikely based on the location described.
    
\end{enumerate}

Given these considerations, the most plausible explanation for the lucent streaks in the left lateral chest wall is subcutaneous emphysema. This conclusion aligns with the characteristic appearance of air tracking through the soft tissues.\textbf{\texttt{</think>}} \\
 
\textbf{\textbackslash \textbackslash boxed\{A: Subcutaneous emphysema\}}
\end{AIbox}
    \caption{Example response for a Chest X-Ray multiple-choice question from RexVQA \cite{pal2025rexvqa}.}
    \label{tab:cxr_example}
    
\end{figure*}

\begin{figure*}[h]
    \begin{AIbox}{Dermatology Example}

   \noindent
   \begin{center}
\begin{minipage}[t]{0.2\textwidth}
    \includegraphics[width=\linewidth]{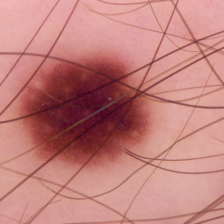}
\end{minipage}%
\hfill
\begin{minipage}[t]{0.75\textwidth}
    \vspace{-3.3cm}
    \footnotesize
    \begin{alltt}
Above is a dermatoscopic image of a pigmented skin lesion. 
Classify the image into one of the following diagnostic categories.
Answer with just one of the options below.
A: Actinic keratoses and intraepithelial carcinoma
B: Basal cell carcinoma
C: Benign keratosis-like lesions
D: Dermatofibroma
E: Melanoma
F: Melanocytic nevi \textcolor{ForestGreen}{(ground truth)}
G: Vascular lesions

Please reason step-by-step, and put your final answer within \textbackslash \textbackslash boxed\{\}.
    \end{alltt}
    
\end{minipage}
\end{center}

\scriptsize

 \footnotesize \textbf{\ours:} \scriptsize \textbf{\texttt{<think>}} To classify this dermatoscopic image, I need to carefully analyze its features and compare them against typical characteristics associated with each diagnostic category. The lesion appears as a uniformly dark brown structure with an overall symmetrical appearance. There are no visible ulcerations, vascular structures, or irregular borders that might suggest malignancy like basal cell carcinoma (B) or melanoma (E). \\

Actinic keratoses and intraepithelial carcinoma (A) typically present as scaly, erythematous patches rather than a pigmented lesion like this one. Vascular lesions (G) usually exhibit red, purple, or blue hues due to blood vessel involvement, which is not evident here. Dermatofibromas (D) often show a central white scar-like area surrounded by a pigment network, but this lesion lacks such features. \\

Benign keratosis-like lesions (C), including seborrheic keratosis, can have varied appearances, including warty or stuck-on textures, but they generally lack the uniform pigmentation seen here. This leaves us with melanocytic nevi (F), which are benign moles characterized by symmetric pigmentation, smooth borders, and uniform coloration—features consistent with this image. \\

Given the symmetry, even distribution of pigmentation, and absence of concerning malignant features, this lesion most closely resembles a melanocytic nevus. \textbf{\texttt{</think>}} \\
 
\textbf{\textbackslash \textbackslash boxed\{F: Melanocytic nevi\}}
\end{AIbox}
    \caption{Example response for a multimodal classification question based on a dermatoscopic image.}
    \label{tab:dermatology_example}
    
\end{figure*}
\begin{figure*}[h]
    \begin{AIbox}{Fundus Example}

   \noindent
   \begin{center}
\begin{minipage}[t]{0.2\textwidth}
    \includegraphics[width=\linewidth]{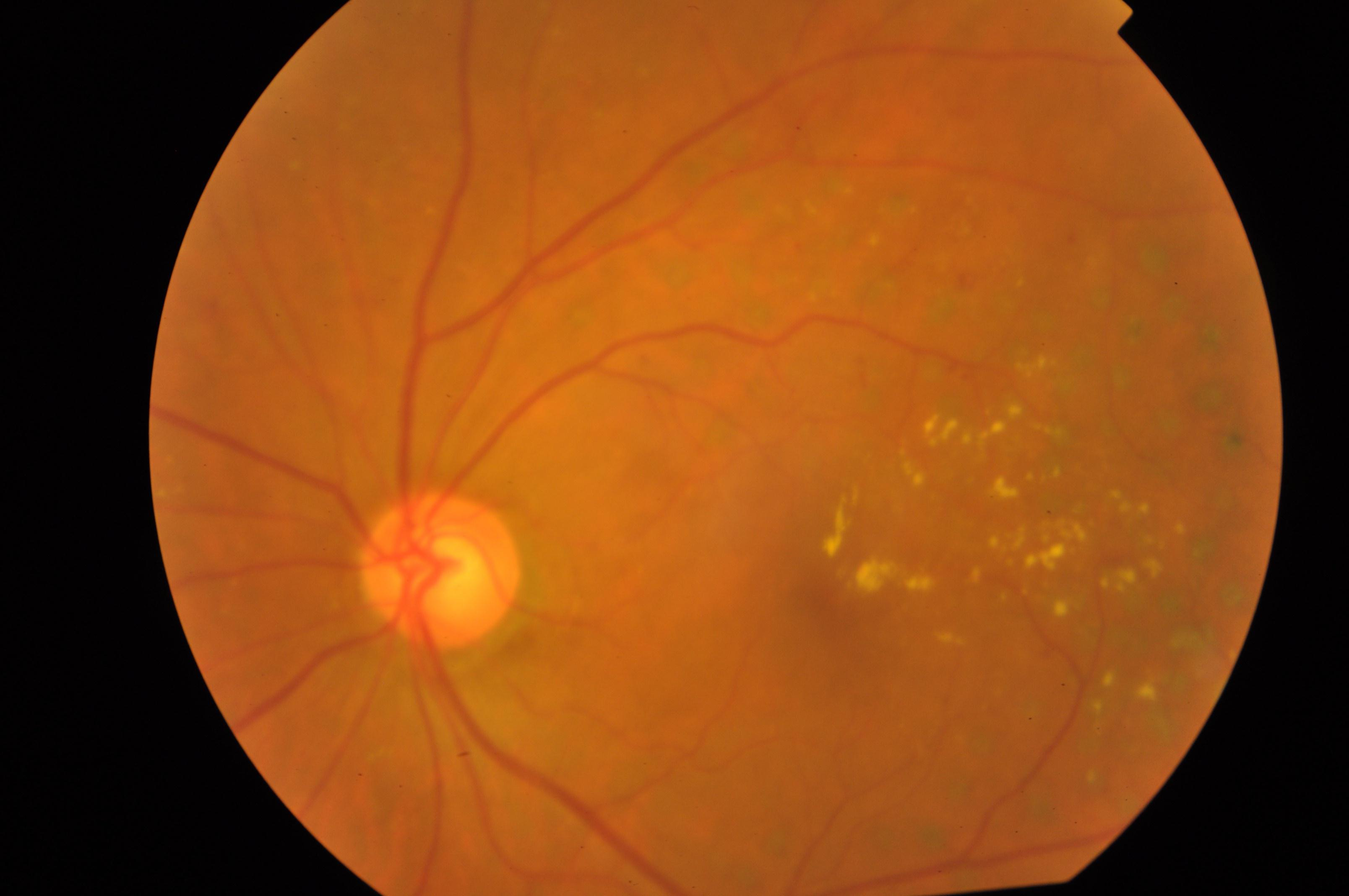}
\end{minipage}%
\hfill
\begin{minipage}[t]{0.75\textwidth}
    \vspace{-2.1cm}
    \footnotesize
    \begin{alltt}
Grade the diabetic retinopathy of the patient in the retinal image on the
Davis Scale of Diabetic Retinopathy. 
A: No DR
B: Mild DR
C: Moderate DR \textcolor{ForestGreen}{(ground truth)}
D: Severe DR
E: Proliferative DR

Please reason step-by-step, and put your final answer within \textbackslash \textbackslash boxed\{\}.
    \end{alltt}
    
\end{minipage}
\end{center}

\scriptsize

 \footnotesize \textbf{\ours:} \scriptsize \textbf{\texttt{<think>}} To assess the diabetic retinopathy grade of the patient using the Davis Scale, I must carefully analyze the retinal image provided. The Davis Scale categorizes diabetic retinopathy into five stages: No DR, Mild DR, Moderate DR, Severe DR, and Proliferative DR. \\

Looking at the retinal image, I notice several key features. The optic disc appears normal, and the retinal vasculature is visible. However, there are yellowish-white spots scattered across the retina, particularly in the area away from the optic disc. These spots are likely hard exudates, which are lipid residues from leaky blood vessels. The presence of hard exudates suggests some level of diabetic retinopathy. \\

There are no signs of neovascularization, which would indicate Proliferative DR. Additionally, I do not observe extensive hemorrhages, venous beading, or intraretinal microvascular abnormalities (IRMA), which are hallmarks of Severe DR. The hard exudates and possible mild retinal changes suggest the condition is less severe than Severe DR. \\

To differentiate between Mild and Moderate DR, I need to consider the extent of the abnormalities. Mild DR is characterized by microaneurysms only, while Moderate DR includes additional features such as hard exudates, more prominent hemorrhages, or cotton-wool spots. The presence of hard exudates in this image suggests the condition is beyond Mild DR and falls into the Moderate DR category. \\

After careful consideration of the retinal features and the Davis Scale criteria, I conclude that the patient has Moderate DR. \textbf{\texttt{</think>}} \\
 
\textbf{\textbackslash \textbackslash boxed\{C: Moderate DR\}}
\end{AIbox}
    \caption{Example response for a multimodal classification question based on a fundus image.}
    \label{tab:fundus_example}
    
\end{figure*}
\begin{figure*}[h]
    \begin{AIbox}{MedQA Example}

   \noindent

\scriptsize
\begin{alltt}
A man is brought into the emergency department by the police department. The officer state that the man has been
arrested multiple times for public alcohol intoxication, but recently became homeless. On exam, the man is
behaving erratically. His vitals are all within normal limits. He appears confused and has a slurred speech.
On gait exam, the patient is ataxic and cannot stand without support for more than a few seconds. Labs return
with the following values: Na 140, K 4, Cl 106, BUN 8, Cr 2. His ABG has pH 7.3, PaCO2 13mm, PaO2 130mm, HCO3 7.
His urinalysis is shown in Figure 1. Blood salicylate levels return as normal. While you await other diagnostic
tests, which of the following should be administered next to treat this patient?
Answer with just one of the options below.
A. Ethanol
B. Naltrexone
C. Naloxone
D. Fomepizole \textcolor{ForestGreen}{(ground truth)}

\end{alltt}

 \footnotesize \textbf{\ours:} \scriptsize \textbf{\texttt{<think>}} Okay, let's try to figure out this case. So, the patient is a man with a history of public intoxication arrests and is now homeless. He's brought in by police, acting erratically, confused, slurred speech, and ataxic. Vitals are normal. Labs show normal sodium, potassium, chloride, BUN, creatinine. ABG is pH 7.3 (which is acidic), PaCO2 13 mmHg (low, so respiratory alkalosis?), HCO3 7 (very low, metabolic acidosis). Wait, pH is 7.3, which is acidotic. So combining the low HCO3 and low PaCO2. Hmm, that's a mixed acid-base disorder? Because in metabolic acidosis, the body compensates by hyperventilating to lower PaCO2. Let me check the anion gap. Na - (Cl + HCO3) = 140 - (106 +7) = 140 -113 = 27. That's a high anion gap metabolic acidosis. Anion gap of 27. So the patient has a high anion gap metabolic acidosis. The ABG pH is 7.3, which is acidotic. The PaCO2 is 13, which is low. So the expected compensation for metabolic acidosis would be using Winter's formula: expected PaCO2 = (1.5 x HCO3) + 8 ± 2. Let's compute that. HCO3 is 7, so 1.5*7 = 10.5 +8 = 18.5. So expected PaCO2 should be around 18.5 ± 2. The patient's PaCO2 is 13, which is lower than expected. That suggests there's an additional respiratory alkalosis component. So maybe a mixed high anion gap metabolic acidosis and respiratory alkalosis? But why the respiratory alkalosis? Could be due to hyperventilation from the acidosis, but the PaCO2 is lower than expected. Hmm. Alternatively, maybe there's another process causing the respiratory alkalosis, like salicylate poisoning? Wait, but salicylate levels are normal. The urinalysis is shown in Figure 1, but since we can't see it, maybe it's something else. The patient is homeless, so maybe alcohol-related issues? But the blood ethanol isn't mentioned here. The options given are ethanol, naltrexone, naloxone, fomepizole. \\

The high anion gap metabolic acidosis could be due to several causes: methanol, ethylene glycol, diabetic ketoacidosis, lactic acidosis, etc. Since the patient is homeless and has a history of intoxication, maybe he ingested something toxic. The options include fomepizole, which is an antidote for methanol and ethylene glycol poisoning. Ethanol is also used as an antidote for those because it competes for the enzyme alcohol dehydrogenase, preventing the conversion of the toxic alcohol into its toxic metabolites. Naltrexone is for alcohol dependence, and naloxone is for opioid overdose. \\

Given the high anion gap metabolic acidosis, and possible toxic alcohol ingestion, fomepizole or ethanol would be the antidotes. Since blood salicylate levels are normal, that's ruled out. So between ethanol and fomepizole. Fomepizole is preferred over ethanol because it's easier to administer and has fewer side effects. So the answer might be D. Fomepizole. But wait, the ABG shows a respiratory alkalosis as well. Could that be a clue? In ethylene glycol poisoning, you can have an initial respiratory alkalosis due to stimulation of the respiratory center, followed by metabolic acidosis. Similarly, methanol can cause similar effects. So if the patient has both high anion gap acidosis and respiratory alkalosis, maybe it's ethylene glycol or methanol. So the treatment would be fomepizole. The other options don't fit. Naltrexone is for alcohol dependence maintenance, not acute intoxication. Naloxone is for opioids. Ethanol could be used as an antidote, but fomepizole is more appropriate. So the answer is D.\textbf{\texttt{</think>}} \\

The patient presents with a high anion gap metabolic acidosis (anion gap = 27) and a mixed acid-base disorder (pH 7.3, PaCO2 13 mmHg, HCO3 7). The low HCO3 and elevated PaCO2 suggest metabolic acidosis with respiratory compensation, but the PaCO2 is lower than expected, indicating a superimposed respiratory alkalosis. This pattern is consistent with toxic alcohol ingestion (e.g., ethylene glycol or methanol), which can cause an initial respiratory alkalosis due to direct stimulation of the respiratory center, followed by metabolic acidosis from toxic metabolites. The patient’s history of homelessness and prior alcohol-related arrests raises suspicion for accidental or intentional ingestion of toxic alcohols. \\

Given the normal salicylate levels and lack of opioid involvement, naloxone (C) is not indicated. Naltrexone (B) is used for alcohol dependence management, not acute toxicity. Ethanol (A) could theoretically compete with toxic alcohols for alcohol dehydrogenase but is less favorable due to side effects and monitoring challenges. **Fomepizole (D)**, a competitive inhibitor of alcohol dehydrogenase, is the preferred antidote for toxic alcohol poisoning, as it prevents the formation of toxic metabolites while being safer and easier to administer than ethanol. 

**Answer:** \textbf{\textbackslash \textbackslash boxed\{D\}}
\end{AIbox}
    \caption{Example of a detailed reasoning trace on a challenging text-only MedQA question.}
    \label{tab:medqa_example}
\end{figure*}  

\renewcommand{\thefigure}{S3-\arabic{figure}}
\setcounter{figure}{0}

\begin{figure*}[h]
    \begin{AIbox}{MedXpertQA Example}

   \noindent
\begin{center}
\begin{minipage}[t]{0.2\textwidth}
    \includegraphics[width=\linewidth]{case_studies/mri.png}
\end{minipage}%
\hfill
\begin{minipage}[t]{0.75\textwidth}
    \vspace{-4.3cm}
    \footnotesize
    \begin{alltt}
This patient was in the recovery phase of diabetic ketoacidosis (DKA).
What is the diagnosis?

Answer Choices:
(A) Basilar artery thrombosis
(B) Osmotic demyelination syndrome \textcolor{ForestGreen}{(ground truth)}
(C) Cobalamin deficiency
(D) Vermian atrophy
(E) Cerebral edema

Please reason step-by-step, and put your final answer within \textbackslash \textbackslash boxed\{\}.
    \end{alltt}
    
\end{minipage}
\end{center}

 \footnotesize \textbf{\ours:} \scriptsize \textbf{\texttt{<think>}} Okay, let's try to figure out this question. The patient is in the recovery phase of diabetic ketoacidosis (DKA), and we need to determine the diagnosis from the given options. First, I should recall what DKA is and its possible complications, especially during recovery. \\

DKA is a serious complication of diabetes where the body starts breaking down fat at a rapid rate, leading to acidic ketones in the blood. It's usually treated with insulin and fluids, but sometimes there are complications during recovery. \\

Now, looking at the MRI image provided, it shows a T2-weighted axial view of the brain. I need to analyze the findings here. The image seems to show symmetrical hyperintensities in the pons, which might be relevant. Hyperintensities on T2 could indicate edema, demyelination, or other pathological processes. \\

Let me consider each option:

A. Basilar artery thrombosis: This would present with acute stroke symptoms. On MRI, you might see restricted diffusion in the affected area. However, the image shows bilateral symmetrical involvement, which is less typical for a thrombosis. Also, basilar artery issues often present with more acute deficits, not necessarily in the recovery phase of DKA. \\

B. Osmotic demyelination syndrome (ODS): Also known as central pontine myelinolysis. This is a known complication of rapid correction of hyponatremia or other metabolic disturbances. DKA treatment involves aggressive fluid administration, which can lead to rapid shifts in osmolality. ODS typically affects the pons, causing symmetrical lesions visible on MRI as hyperintensities on T2/FLAIR. The clinical scenario fits because the patient is recovering from DKA, which involves fluid and electrolyte shifts. \\

C. Cobalamin deficiency: This leads to subacute combined degeneration of the spinal cord, not typically pontine lesions. It's associated with megaloblastic anemia and neurological symptoms like ataxia, but the MRI findings here don't support this diagnosis. \\

D. Vermian atrophy: This refers to shrinkage of the cerebellar vermis, often seen in chronic conditions like alcoholism or certain neurodegenerative diseases. It's not related to DKA recovery and wouldn't present with acute changes in the pons. \\

E. Cerebral edema: While cerebral edema can occur in DKA, especially during treatment, it usually presents more diffusely and acutely. The MRI here shows localized lesions in the pons rather than diffuse swelling. \\

Putting this together, the key points are the recovery phase of DKA, the MRI showing symmetrical pontine lesions, and the association with osmotic shifts. Osmotic demyelination syndrome (B) is the most likely diagnosis. Basilar thrombosis doesn't fit the imaging pattern or the clinical context. Cobalamin deficiency and vermian atrophy don't match the presentation. Cerebral edema would be more widespread and acute. \textbf{\texttt{</think>}} \\

The patient is in the recovery phase of diabetic ketoacidosis (DKA), a condition known to cause significant metabolic and osmotic disturbances. During treatment, rapid correction of electrolyte imbalances—particularly sodium levels—can lead to **osmotic demyelination syndrome (ODS)**, previously termed central pontine myelinolysis. This condition is characterized by damage to the myelin sheaths in the central nervous system, most commonly affecting the pons, due to rapid shifts in osmolality.

---

\#\#\# Key Clinical and Radiological Features Supporting ODS:
- **Clinical Context**: DKA management often involves aggressive rehydration and insulin therapy, which can lead to abrupt changes in serum osmolality. This creates a high-risk environment for ODS.
- **Imaging Findings**: The MRI demonstrates symmetrical hyperintensities in the pons on T2-weighted imaging, which is classic for ODS. These lesions are typically bilateral and localized to the upper pons, consistent with the pathophysiology of demyelination following osmotic stress.
- **Timing**: ODS typically manifests 2–4 days after the correction of hyponatremia or other metabolic derangements, aligning with the patient being in the **recovery phase** of DKA.

---

\#\#\# Evaluation of Other Options:
- **A. Basilar artery thrombosis**: This would present with acute neurological deficits and restricted diffusion on MRI, not the symmetrical T2 hyperintensities seen here.
- **C. Cobalamin deficiency**: This causes spinal cord pathology (e.g., subacute combined degeneration), not pontine lesions.
- **D. Vermian atrophy**: This is a chronic, degenerative condition affecting the cerebellum, not an acute process related to DKA.
- **E. Cerebral edema**: While possible in DKA, it would present with diffuse brain swelling and loss of gray-white differentiation, not localized pontine changes.

---

\#\#\# Conclusion:
Given the clinical context of DKA recovery and the characteristic MRI findings, **osmotic demyelination syndrome** is the most likely diagnosis.

\$\$

\textbf{\textbackslash \textbackslash boxed\{B\}}

\$\$
    \end{AIbox}
    \caption{Example of a detailed reasoning trace on a challenging MedXpertQA question. \ours displays longer reasoning traces for challenging questions.}

    \label{tab:medxpertqa_example}
\end{figure*}
\begin{figure*}[h]
    \begin{AIbox}{PMC-VQA Example}

   \noindent
   \begin{center}
\begin{minipage}[t]{0.2\textwidth}
    \includegraphics[width=\linewidth]{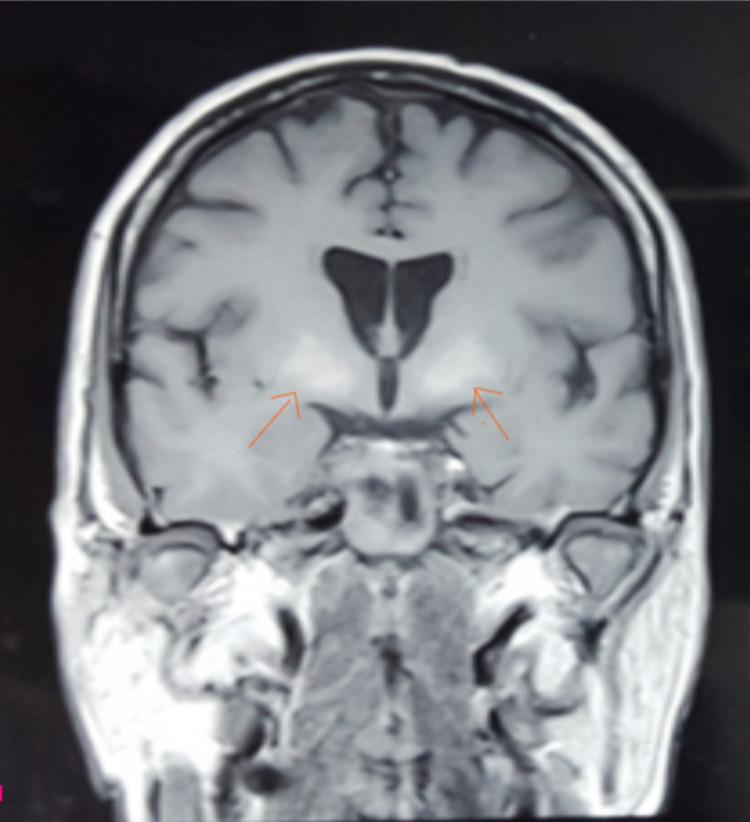}
\end{minipage}%
\hfill
\begin{minipage}[t]{0.75\textwidth}
    \vspace{-3.3cm}
    \footnotesize
    \begin{alltt}
What is the name of the medical imaging technique used in this case?
A: X-ray
B: Magnetic resonance imaging \textcolor{ForestGreen}{(ground truth)}
C: Computed tomography
D: Ultrasound

Please reason step-by-step, and put your final answer within \textbackslash \textbackslash boxed\{\}.
    \end{alltt}
    
\end{minipage}
\end{center}

\scriptsize

 \footnotesize \textbf{\ours:} \scriptsize \textbf{\texttt{<think>}} Okay, let's analyze the question and the provided image. The image shows a detailed view of the brain in a coronal section, with clear visualization of soft tissues, ventricles, and other intracranial structures. This level of detail, especially for soft tissues, is characteristic of Magnetic Resonance Imaging (MRI). X-rays (option A) are primarily used for imaging bones and do not provide this level of detail for soft tissues. Computed Tomography (CT, option C) can show soft tissues but is less detailed compared to MRI and often involves the use of ionizing radiation. Ultrasound (option D) is not suitable for imaging the brain in this manner due to the skull obstructing sound waves. Therefore, the medical imaging technique used in this case is Magnetic Resonance Imaging (MRI). \textbf{\texttt{</think>}} \\
 
\textbf{\textbackslash \textbackslash boxed\{B: Magnetic resonance imaging\}}
\end{AIbox}
    \caption{Example reasoning trace on a question from PMC-VQA. \ours displays shorter reasoning length for easier questions.}
    \label{tab:pmc_vqa_example}
\end{figure*}

\begin{itemize}
    \item \textbf{Task Versatility: } \ours is capable of various clinical reasoning tasks by simply changing the prompt. As shown in Figure \ref{tab:medication_assessment} and Figure \ref{tab:cxr_example}, \ours can provide advice on medication assessment or provide image reports.
    \item \textbf{Modality Versatility: } \ours is able to answer queries about images from many different imaging modalities, such as Pathology (Figure \ref{tab:bcss_example}), Chest X-Ray (Figure \ref{tab:cxr_example}), Dermatology (Figure \ref{tab:dermatology_example}), Fundus (Figure \ref{tab:fundus_example}), and Text-Only (Figure \ref{tab:medqa_example}).
    \item \textbf{Task-Aware Thinking: } \ours displays a propensity to adapt its reasoning trace length depending on the downstream task. For more challenging tasks such as MedXpertQA, \ours has longer average reasoning traces (Figure \ref{tab:medxpertqa_example}). However, simpler queries such as those from PMC-VQA result in shorter chain of thought length (Figure \ref{tab:pmc_vqa_example}).
\end{itemize}

\section{Question Sources for Evaluation Datasets}
To comprehensively evaluate multimodal medical reasoning, we considered three benchmark categories. \textbf{Text-Only} tasks evaluate the model's ability to apply its knowledge to solve multi-step clinical reasoning tasks without visual cues. \textbf{Multimodal Reasoning} benchmarks test the model's capability to combine its text-only reasoning capabilities with visual evidence. \textbf{Multimodal Classification} tasks measure how well models can use perceptual grounding to extract clinical relevant features and arrive at diagnoses. Together, these categories provide comprehensive assessment of reasoning, modality integration, and perception, each critical for building a trustworthy multimodal reasoning model. We provide more details about the evaluation datasets we used for each task category below.

\label{sec:dataset_information}

\subsection*{Text-Only}
\begin{itemize}
\item \textbf{MedQA} \cite{jin2021disease} is a large-scale medical multiple-choice QA dataset drawn from U.S. medical licensing exams. We use the english subset which contains 12,723 English questions, with 10,178 for training, 1,272 for validation, and 1,273 for testing. Each question is text-only and has four answer options (one correct), making it a 4-way multiple-choice classification task. There is a corresponding 5-option version of the dataset, which we used to curate our training data. We combine the training and validation splits of the 5-option version to use as SFT data, and evaluate on the 4 option version test split.
\item \textbf{MMLU-PRO} (Health Subset) \cite{wang2024mmlu} is a challenging multiple-choice benchmark which builds off of the original MMLU benchmark \cite{hendrycks2020measuring} by expanding the number of multiple choice options to 10 per question. We evaluate on the health subset which contains 818 questions.

\item \textbf{MedMCQA} \cite{pal2022medmcqa} is a 4-option multiple-choice benchmark of real-world medical exams containing 182,822, 4,183, and 6,150 questions in the train, val, and test splits respectively.
\item \textbf{MedXpertQA} (text) \cite{zuo2025medxpertqa} is a comprehensive reasoning benchmark consisting of 17 medical specialities and 11 body systems. In the text-only portion of the dataset contain 2455 questions, each question has 10 multiple choice options. We only use this dataset for evaluation to test the generalization ability of \ours.
\item \textbf{HeadQA} \cite{vilares-gomez-rodriguez-2019-head} contains questions from medical exams designed to evaluate readiness to access specialized portions of the Spanish medical system. There are 2657, 1366, and 2742 questions in the train, val, and test splits respectively. We merge the train and val splits for distillation and supervised finetuning, and evaluate on the unseen test split.
\end{itemize}

\subsection*{Multimodal Reasoning}
\begin{itemize}
\item \textbf{PMC-VQA} \cite{zhang2023pmc} is a large-scale medical VQA benchmark consisting of over 227,000 image-question pairs and 149,000 unique images. The authors also provide a manually verified test split consisting of 2000 image-question pairs. We evaluate on this clean test split and carefully remove any questions from our training data with image overlap.
\item \textbf{MedXpertQA} (multimodal) \cite{zuo2025medxpertqa} contains 2000 questions related to advanced multimodal reasoning about various medical specialties and body systems. Each question is multiple choice and has 5 options.
\item \textbf{MMMU-PRO} \cite{yue2024mmmu} is a widely used college-exam level multimodal reasoning benchmark. Each question is multiple choice and may refer to one or more images required to correctly arrive at the answer. We evaluate on the medical subset of the pro version of the dataset, which augments the answer choices to include up to 10 options. There are a total of 286 questions in the medical split.
\item \textbf{NEJM Image Challenge} \cite{nejm_image_challenge} is a weekly challenge hosted by the New England Journal of Medicine in which participants are tasked to perform differential diagnoses to assess patient condition based on their case report summary and accompanying images. We follow prior work which scraped the questions from past weeks resulting in a total of 947 5-option multiple choice questions about multimodal patient differential diagnosis.
\end{itemize}

\subsection*{Multimodal Classification}
\begin{itemize}
\item \textbf{Brain Tumor Classification} \cite{bhuvaji2020brain} is a MRI image dataset consisting of T1-Weighted images of brain tumors. The objective is to classify the tumor into one of 4 categories: Glioma, Meningioma, Pituitary, or no tumor. We follow the code in the CLIMB \cite{dai2025climb} codebase to obtain a train and test split of 2,870 and 394 samples respectively.
\item \textbf{Coronahack} \cite{nasir2023multi} is a Chest X-ray dataset in which models must predict if the patient has Bacterial Pneumonia, Viral Pneumonia, or is normal. We follow the code in the CLIMB codebase to obtain a train and test split of 5,284 and 624 records respectively.
\item \textbf{Aptos} \cite{aptos2019} is a fundus imaging dataset collected by the Asia Pacific Tele Ophthalmology Society Symposium. The task is to classify the diabetic retinopathy rating of a patient's fundus image into one of 5 categories: No DR, Mild, Moderate, Severe, or Proliferative DR. Following the code in the CLIMB codebase, we obtain the same train test split of 2,929 and 733 image-question pairs respectively.
\item \textbf{MESSIDOR-2} \cite{decenciere2014feedback} Similar to Aptos, the MESSIDOR-2 dataset contains fundus images labelled with their corresponding diabetic retinopathy grades. We follow the CLIMB codebase to obtain a train-test split of 1,394 and 350 records.
\item \textbf{BCSS} \cite{amgad2019structured} is a crowd-sourced dataset of pathology slides of breast tissue. The task is to classify the tissue into one of 4 categories: Tumor, Stroma, Inflammatory, or Necrosis. We follow the CLIMB codebase to obtain our train-test split.

\end{itemize}
\begin{figure}[h]
    \begin{AIbox}{GPT-4o Distillation Prompt}
    \parbox[t]{\linewidth}{

     \scriptsize \begin{alltt}
    You should provide your detailed thoughts within <think> </think> tags, always making sure to reflect and think about your response, then answer with just one of the options below within <answer> </answer> tags. Your response should carefully consider the options and output a very long chain of thought. (For example, if the question is \\'Is the earth flat?\\A: Yes \\B: No', you should answer with <think>Okay, let's tackle this question about whether the Earth is flat or not. The idea that the Earth is flat may feel intuitive because our everyday experience seems flat, but overwhelming evidence shows it's a sphere. First, astronomical observations reveal that stars rotate differently in the northern and southern hemispheres: Polaris is visible up north but not down south, which only makes sense on a curved surface. Wait, could that be due to perspective? No, let me double check; this change in visible stars directly correlates with latitude, which wouldn’t happen on a flat plane. Ships also disappear bottom-first over the horizon; wait, maybe that’s just perspective? But even with a telescope, the hull stays hidden, confirming it’s curvature, not optics. Then there's air travel: planes follow great-circle routes, which look curved on flat maps but are the shortest path on a globe. Let me double check. Yes, for example, New York to Tokyo arcs over Alaska only because the Earth is round. During lunar eclipses, Earth always casts a round shadow on the Moon. Wait, could a flat disc do that? Only from one angle; a sphere is the only shape that does this consistently. And what about space photos? Are they fake? Let me double check. No, not just NASA, but independent agencies and private companies would all have to be complicit, and their satellite systems rely on spherical Earth physics to work, including GPS. Time zones also show curvature; when it’s day in Tokyo, it’s night in New York. Wait, could the Sun just be a spotlight above a flat Earth? That fails too; we’d see the Sun all the time just dimmer, not dipping below the horizon. Also, engineers designing long bridges or tunnels adjust for curvature, and GPS satellites wouldn’t function without spherical Earth modeling. Let me double check—yes, geodetic surveying and orbital mechanics prove it. So from ancient Greek reasoning to modern engineering and spaceflight, every independent line of evidence confirms the Earth is not flat, but round.</think> <answer>B: No</answer>). Here is the question: \{\{ question \}\}
    \end{alltt}}

    \end{AIbox}
    \caption{Prompt used with GPT-4o to curate reasoning traces for multimodal tasks.}

    \label{tab:gpt_4o_prompt}
\end{figure}
\begin{figure}[h]
    \begin{AIbox}{DeepSeek-R1 Distillation Prompt}
    \parbox[t]{\linewidth}{
    
     \scriptsize \begin{alltt}
    \{\{question\}\}
    
    
    Put your final answer letter within <answer></answer> tags.
    
    \end{alltt}}

    \end{AIbox}
    \caption{Prompt used with DeepSeek-R1 to curate reasoning traces for text-only tasks.}

    \label{tab:deepseek_prompt}
\end{figure}

\begin{figure}[h]
    \begin{AIbox}{\ours Evaluation Prompt}
    \parbox[t]{\linewidth}{
    
     \scriptsize \begin{alltt}
\{\{question\}\}


Please reason step-by-step, and put your final answer within \textbackslash \textbackslash boxed\{\}.
    \end{alltt}}

    \end{AIbox}
    \caption{Prompt used to evaluate \ours on multiple choice tasks.}

    \label{tab:omr_prompt}
\end{figure}

\begin{figure}[h]
    \begin{AIbox}{Huatuo-GPT-Vision Evaluation Prompt}
    \parbox[t]{\linewidth}{

     \scriptsize \begin{alltt}
\{\{question\}\}


Answer with the option’s letter from the given choices directly.
    \end{alltt}}

    \end{AIbox}
    \caption{Prompt used to evaluate \texttt{Huatuo-GPT-Vision} on multiple choice tasks. Since we found the model struggles to perform long chain of thought, we copy the template from their work directly.}

    \label{tab:huatuo_prompt}
\end{figure}

\begin{figure}[h]
    \begin{AIbox}{LingShu Evaluation Prompt}
    \parbox[t]{\linewidth}{

     \scriptsize \begin{alltt}
Question: \{\{question\}\}

Answer with the option's letter from the given choices and put the letter in one "\textbackslash \textbackslash boxed\{\}"
    \end{alltt}}

    \end{AIbox}
    \caption{Prompt used to evaluate \texttt{LingShu-7B} on multiple choice tasks. We use the default reasoning prompt int their MedEvalToolkit without modifications, reproduced here for convenience.}

    \label{tab:lingshu_prompt}
\end{figure}

\begin{figure}[h]
    \begin{AIbox}{MedVLThinker Evaluation Prompt}
    \parbox[t]{\linewidth}{

     \scriptsize \begin{alltt}
You will solve a problem/request. You should provide your thoughts within <think> </think> tags before providing the answer.

Write your final answer within <answer> </answer> tags.

\{\{question\}\}
    \end{alltt}}

    \end{AIbox}
    \caption{Prompt used to evaluate \texttt{MedVLThinker} on multiple choice tasks. We use the prompt in the appendix of their work, reproduced here for convenience.}

    \label{tab:medvlthinker_prompt}
\end{figure}
\section{Prompts}

\renewcommand{\thefigure}{S3-\arabic{figure}}
\paragraph{Distillation Prompts} Figure \ref{tab:gpt_4o_prompt} and Figure \ref{tab:deepseek_prompt}
 show the distillation prompt used to prompt GPT-4o and DeepSeek-R1 for multimodal and text-only tasks respectively.
\renewcommand{\thefigure}{S4-\arabic{figure}}
\paragraph{Evaluation Prompts} To ensure fair comparison, we did our best to evaluate all models using their suggested prompts. For transparency, we show the prompts used for each model in this section in Figures \ref{tab:omr_prompt} to \ref{tab:medvlthinker_prompt}.

\label{sec:prompts}

\end{document}